\documentclass{article}
\usepackage{arxiv}

\usepackage{hyperref}
\usepackage{bm}% allocate bold fonts
\usepackage{cite}
\usepackage{graphics,fancyhdr,graphicx,subfigure}
\usepackage{subfigure}
\usepackage{amsthm,amsmath,amsfonts,latexsym,amssymb}
\usepackage{lineno}
\usepackage{diagbox}
\usepackage{algorithm}
\PassOptionsToPackage{noend}{algpseudocode}
\usepackage{algpseudocode}
\usepackage{listings}
\usepackage[table]{xcolor}
\usepackage{lscape}
\usepackage{comment}
\usepackage{tabularx}
\usepackage{graphicx}%to add images
\usepackage{tabularx}
\usepackage{nomencl}%for adding symbols page
\usepackage{array}%to add table
\usepackage{caption}%add caption for the table and list it in list of tables
\usepackage{breqn}
\usepackage{lineno}
\usepackage{mathtools}
\usepackage{subfigure}
\usepackage{caption}
\usepackage{lmodern}
\usepackage{calligra}
\usepackage{xspace}
\usepackage[utf8]{inputenc}%skiplength
\usepackage{enumerate}%list
\usepackage[english]{babel}
\def\equationautorefname~#1\null{%
  Eq.~(#1)\null
  }
\def\subfigureautorefname~#1\null{%
  Fig.~#1\null
}

\definecolor{listinggray}{gray}{0.9}
\definecolor{lbcolor}{rgb}{0.9,0.9,0.9}
\definecolor{Darkgreen}{RGB}{0,100,0}

\title{Koopman operator for time-dependent reliability analysis \thanks{https://www.csccm.in/}}

\author{ \hspace{1mm}Navaneeth~N.\\
	Department of Applied Mechanics\\
	Indian Institute of Technology (IIT) Delhi\\
	Hauz Khas - 110 016, New Delhi, India \\
	\texttt{navaneeth.n@am.iitd.ac.in} \\
	\And
	\hspace{1mm}Souvik~Chakraborty \\
	Department of Applied Mechanics\\
	Indian Institute of Technology (IIT) Delhi\\
	Hauz Khas - 110 016, New Delhi, India \\
	\texttt{souvik@am.iitd.ac.in} \\
}

\renewcommand{\shorttitle}{Koopman operator for time-dependent reliability analysis}

\hypersetup{
pdftitle={Koopman operator for time-dependent reliability analysis},
pdfsubject={q-bio.NC, q-bio.QM},
pdfauthor={Navaneeth N., Souvik Chakraborty},
pdfkeywords={time-dependent reliability, Koopman operator, First passage failure time},
}

\begin{document}
\maketitle

\begin{abstract}
Time-dependent structural reliability analysis of nonlinear dynamical systems is non-trivial; subsequently, scope of most of the structural reliability analysis methods is limited to time-independent reliability analysis only. In this work, we propose a Koopman operator based approach for time-dependent reliability analysis of nonlinear dynamical systems. Since the Koopman representations can transform any nonlinear dynamical system into a linear dynamical system, the time evolution of dynamical systems can be obtained by Koopman operators seamlessly regardless of the nonlinear or chaotic behavior. Despite the fact that the Koopman theory has been in vogue a long time back, identifying intrinsic coordinates is a challenging task; to address this, we propose an end-to-end deep learning architecture that learns the Koopman observables and then use it for time marching the dynamical response. Unlike purely data-driven approaches, the proposed approach is robust even in the presence of uncertainties; this renders the proposed approach suitable for time-dependent reliability analysis. We propose two architectures; one suitable for time-dependent reliability analysis when the system is subjected to random initial condition and the other suitable when the underlying system have uncertainties in system parameters. The proposed approach is robust and generalizes to unseen environment (out-of-distribution prediction).
Efficacy of the proposed approached is illustrated using three numerical examples. Results obtained indicate supremacy of the proposed approach as compared to purely data-driven auto-regressive neural network and long-short term memory network. 
\end{abstract}

% keywords can be removed

\section{Introduction}
\label{sec:intro}
% Most of the physical systems exhibit significant deviations from the desired response due to random fluctuations in operational, environmental conditions and varying system properties. However, it is required to ensure that the system meets certain performance standards regardless of the randomness associated system. When all output measurements of the system are stochastic, it is essential to quantify the uncertainty and hence reliability analysis methods are employed.
Structural reliability can be defined as the ability of the structure to perform desirable tasks over a specified time under operational and environmental conditions. In general, reliability analyses are categorised into two based on the time-dependency of the performance function; (a) Time-independent reliability analysis and (b) Time-dependent reliability analysis.  For the case of time-independent reliability analysis, the reliability and failure probability of a system remains unchanged over time. On the other hand, for time-dependent reliability analysis, the failure probability of the system varies with time as the system response measurements contributing to the limit state function change with time. Note that reliability by definition is time-dependent, and time-independence is only an assumption.
\par
For determining time-independent reliability analysis, various established methods are available in the literature. Popular examples of such methods include the First Order Reliability Method (FORM) \cite{hu2015first}, Second-Order Reliability Method (SORM) \cite{zhang2010second}, Response Surface Methods (RSM) \cite{faravelli1989response, alibrandi2015new}, Importance Sampling Method (ISM) \cite{melchers1989importance}, Monte-Carlo Simulations (MCS)\cite{thakur1978monte, rubinstein2016simulation} and surrogate model based approaches \cite{chakraborty2017efficient, ghosh2018support, bilionis2012multi, roy2019support, sudret2008global, garg2022assessment}. Nevertheless, these time-independent approaches can not be employed for most of the realistic structures as the system responses are dynamic in nature and system parameters vary with time. This necessitates the development of efficient reliability methods for time-dependent systems. Generally, time-dependent reliability analysis problems are solved by converting it into a time-independent reliability analysis problem \cite{wu2021time, mori1993time, zhang2017time}; however, such a setup is restricted in the sense that one cannot compute first passage failure time. It is to be noted that first passage failure time is extremely important particularly in case of structural health monitoring \cite{aditya2008sensitivity, friswell2010structural, bhowmik2019first, lenjani2019hierarchical} and residual life estimation \cite{wang2014residual, herzog2009machine}.
Time-dependent reliability analysis methods have recently received much attention due to rising needs in practical problems. In this paper also, our focus is on time-dependent reliability analysis only and hence, the discussion hereafter is restricted to the same.
% Here, we are more concerned with the time-dependent probability of failure because it provides us with the possibility of a product performing its intended function over its service time or the system meets certain performance standards throughout its life cycle.
\par
Several time-dependent reliability analysis methods have been proposed in various studies, however, the extreme value distribution method \cite{hu2013sampling, chen2007extreme} is one of the simplest approaches employed in computing reliability of time-dependent systems. The method can be used when the distribution of the extreme value of the limit-state function over the given period is known prior. Nevertheless, in the majority of the cases, the distribution of extreme value may not be available in an explicit form. Another prominent approach is the out-crossing rate method \cite{sudret2008analytical,hagen1991vector}, where the concept of first passage failure proposed by Rice \cite{rice1944mathematical} is utilized to compute failure rate.  According to the Rice formulation for the outcrossing rate, the mean number of out-crossings per unit time is obtained as the first-time crossing rate or the failure rate. Here the notion of out-crossing is equivalent to the event of up crossing of limit state function or the dynamic response over the permissible threshold. Rice's formulation outperforms other methods in terms of computing efficiency. Further, many improvements of Rice's formulation based methods are proposed by researchers \cite{engelund1995approximations,rackwitz1998computational,lutes2009reliability,zhang2011time}. Hu and Mahadevan \cite{hu2016single} devised a different approach, a surrogate-based approach for time-dependent reliability analysis known as a single-loop Kriging. Another meta-modelling technique proposed by Drignei et al. \cite{drignei2016random} calculates the time-dependent likelihood of failure by defining the stochastic process of output response in a dynamic system. Although several methods exists in the literature, the first-passage technique is the extensively used one to compute the first time out-crossing rate to obtain time-dependent reliability.

%Monte Carlo Simulation (MCS) method
%Koopman literature survey
Analysis of dynamical systems is a key component in time-dependent reliability estimation. While analysis of the linear dynamical system is straightforward, the same cannot be said about the nonlinear dynamical system.  
% Dynamical system evolves with time in such a way that the current state of the system has dependencies on previous time steps. When the system under consideration is linear, the time evolution of it can be easily achieved through general procedures. Nevertheless, in most of the cases, dynamical systems are complex and inherently nonlinear. In that case, a general framework can not be adopted and thus analysis of nonlinear dynamical systems is not straight forward.
Spectral methods, which facilitate the analysis of complex dynamical systems, gained wide popularity in recent times because of the abundance of data. Dynamic Mode Decomposition (DMD) is one such methods, proposed by Schmidt \cite{schmid2010dynamic} where one utilizes the computationally efficient proper Singular Value Decomposition (SVD). Another promising approach is Koopman operator-based approximation \cite{koopman1931hamiltonian}. The unique feature of Koopman representations is that they can transform any nonlinear dynamical system into a linear dynamical system through Koopman observables. While DMD and the Koopman theory were first suggested independently, later a strong connection between DMD and the Koopman theory has been established by Rowley \cite{rowley2009spectral}. Although in theory, Koopman observables simplify the dynamics of any complex dynamical system, infinite-dimensional Koopman coordinates, as postulated, are not practically feasible. Even when coordinates are truncated to finite dimensions, identification of the coordinates is not straightforward.
% However, the emergence of data-driven methods, especially deep learning models, provides for the efficient learning of Koopman coordinates \cite{li2005curse}.
\par
\textit{Deep learning}, in recent times, has been recognized as a potential solution in a broad range of applications including computer vision, natural language processing, Internet of Things (IoT), speech processing, neuroscience, autonomous driving car and so on. Motivated by these developments, the objective of this paper is to develop a data-driven deep learning framework that can learn the Koopman operator and use it for time-dependent reliability analysis. In this paper, two deep learning architectures are based on the Koopman operator. While the first architecture is suitable for problems with the random initial condition, the second one can incorporate uncertainty in system parameters. Some important features of the proposed approach include:
% An Artificial Neural Network (ANN) is an information processing paradigm that is inspired from biological nervous systems.The frameworks are now used for a wide range of computational tasks. Artificial neural network and \textit{deep learning} have been recognized as a potential solution in a broad range of applications including computer vision, natural language processing, Internet of Things (IoT), speech processing, neuroscience, autonomous driving car and so on. 
% Neoc framework.
\begin{itemize}
    \item The proposed framework incorporates Koopman based loss-function during training the deep learning model. This results in better generalization and higher accuracy.
    \item The method is scalable as the dictionaries of Koopman observables are automatically selected.  
    \item When existing spectrum approaches, such as DMD and Extended DMD, are ideally suited to dynamical systems with given initial conditions and system parameters, the proposed Koopman framework effectively predict system states even when the initial conditions and system parameters are uncertain. This is particularly important as the objective is to use the proposed framework for time-dependent reliability analysis.
\end{itemize}

\par
The remainder of the paper is organised as follows. The general problem statement is described in the Section \ref{sec:Problem statement}. The proposed approach is elucidated in the following Section \ref{sec:Proposed approach}. Moreover,  Numerical examples are presented subsequently in Section\ref{sec: Numerical example}. Finally, the conclusion and final notes are provided in Section \ref{sec:Conclusions}.

\section{Problem statement}\label{sec:Problem statement}
For a time-independent system having response $g(\bm{Y})$ and threshold $e_h$, failure probability $P_f$ is given by:
\begin{equation}
    P_{f}=\mathbb P(g(\bm{Y})<e_h),
\end{equation}
where  $\bm{Y}$ is an N-dimensional vector of random variables $\bm{Y}$, $\bm{Y}=\left(Y_{1}, Y_{2},...., Y_{N}\right)$ and $\mathbb P$ is the probability. 
In the case of time-independent systems, output response, $g$ depends only on the random parameters. The response of a time-dependent system, on the other hand, is governed by the random parameters ($\bm Y$) as well as the time (t). Thus, failure probability can be expressed as:
\begin{equation}
    P_{f}(t)=\mathbb P(g(\bm{Y},t)<e_h(t)).
\end{equation}
Here it should be noted that the threshold of the system ($e_h$) may also vary with time.
In the present work, we employ the first passage failure method to evaluate the time-dependent failure probability of dynamical systems, where we firstly focus on the First Time To Failure (FTTF), also known as first passage failure time. FTTF is defined as the time ($\tau$) at which the system response, $g({.})$ exceeds the threshold for the first time. As a result, the failure probability is as follows:
\begin{equation}\label{eq:time_threshold}
  P_{f}=\mathbb P(g(\bm{Y},\tau)<e_h(\tau), \tau\in [t_o,t_s]),
\end{equation}
where $t_0$ and $t_s$ are the initial and final points of the time interval under which the system is considered. Because of uncertainty in $\bm Y$, $\tau$ is also a random variable. The objective here is to compute the Probability Density Function (PDF) of the first passage failure time. 

\section{Proposed approach}\label{sec:Proposed approach}
The data-driven Koopman approach presented in this study is described in detail in this section. Before delving into the specifics of our deep-learning approach for the Koopman operator, we briefly review the basic formulations of dynamical systems and Koopman dynamics. 
\subsection{Basic mathematical formulations}
Here we consider a general dynamical systems of the form
\begin{equation}
\frac{d}{d t} \bm{X}(t)=\mathbf{f}(\bm{X}(t), t ; \bm Y),
\end{equation}
where $\bm{X}$ is the state variables such that $\bm{X}\in \mathbb{R}^{n}$, $t$ is the time and $\alpha$ is the system parameters. On the other hand, discrete dynamical system is represented as
\begin{equation}
    \bm{X}_{k+1}=\mathbf{F}(\bm{X}_k).
\end{equation}
Here, $\bm{x}_k$ and $\bm{x}_{k+1}$ represent state variables of the system at $k^{th}$ and ${k+1}^{th}$ time steps, and $\mathbf{F}$ represents the mapping of the system variables in forward time. Discrete-time dynamics are often generated from the continuous dynamical system through time discrete sampling. Thus, for a given arbitrary time step, the system evaluation of through map can be represented as 
\begin{equation}
    \bm X(t+\Delta t)=\mathbf{F}(\bm X(t_0))=\bm X(t_0)+\int\limits_t^{t + \Delta t} f({\bm X(\tau)}){d\tau}
\end{equation}
For a linear dynamical system, the time evaluation map, $\mathbf F$ becomes a matrix that is characterised by eigenvalues. Thus, for linear dynamical systems, time evaluation can be achieved by the linear mapping matrix seamlessly. However, for nonlinear dynamical systems, the mapping can be arbitrarily complex. So, it is not straightforward to achieve the time evolution of the nonlinear systems. Koopman's theory made it possible to analyze a nonlinear dynamical system through the Koopman representations. According to the theory, any dynamical system can be represented using an infinite-dimensional linear operator acting on a Hilbert space of measurement functions,
\begin{equation}
    \mathcal{K} \psi \left({\bm X}_{i}\right) \triangleq \psi \circ \mathbf{F}\left(\bm{X}_{i}\right)
\end{equation}
where ${\psi}$ is the measurement functions, known as the Koopman observables and $\mathcal{K}$ is the linear operator, known as the Koopman operator. Here, $\psi \circ \mathbf{F}$ represents the action of the composite operator and is given by:
\begin{equation}
 \psi \circ \mathbf{F}\left(\mathbf{X}_{i}\right)= \mathbf F(\psi \left(\bm X_{k}\right))=\psi \left(\bm X_{k+1}\right)
\end{equation}
Once the Koopman representations are obtained, time evaluation can be obtained as:
\begin{equation}
     \psi\left(\bm X_{k+1}\right)=\mathcal{K}^{n}\psi\left(\bm X_{k}\right)
\end{equation}
The Koopman representations simplify the dynamics of complex nonlinear systems. However, as the theory states, lifting the state space variable into the infinite-dimensional observables is not practically feasible. Moreover, finite-dimensional approximation of the observables is also perhaps a challenging task.  
% Coming to the being of data driven approach, in the recent years,  motivated the researchers \cite{takeishi2017learning,lusch2018deep,otto2019linearly,korda2018linear} to use deep learning and machine learning methods identify the finite Koopman representations. Neural networks are capable of representing any arbitrarily complex functions. This persuades us to utilize a deep learning based framework to model Koopman dynamics.
\subsection{Data driven Koopman representation}\label{Koopman_rep}
In this section, we propose two deep learning architectures for learning Koopman representation and time-dependent reliability analysis. The first architecture proposed is useful for systems with uncertainty in the initial conditions. The second architecture, on the other hand, is suitable for systems with uncertainty in the model parameters. Details on the two architectures are furnished below.\par
%-----------------   Rewrite below  -----------------------
Consider $\bm X_k$ to be the state of a dynamical system at the ${k}^{th}$ time step and $\bm X_{k+1}$ to be the state variable at ${k+1}^{th}$ time state. The objective here is to develop a deep learning model $\mathcal M$ that can learn the mapping from $\bm X_k$ to $\bm X_{k+1}$,
\begin{equation}
    \mathcal M: \bm X_k \mapsto \bm X_{k+1}
\end{equation}
The obvious choices here are to either use auto-regressive deep learning models \cite{triebe2019ar} or sequence learning models such as recurrent neural network (RNN) \cite{connor1994recurrent} and long-short term memory (LSTM) \cite{guo2021mechanical}. Literature suggests that both types of models perform reasonably well for time-series modeling; however, the objective here is to perform time-dependent reliability analysis of physical systems and direct application of these models are unlikely to work (illustrated numerically in \autoref{sec: Numerical example}). Therefore, we here propose an auto-regressive model constrained by the Koopman dynamics. To that end, the state vector, $\bm X_k$ is first lifted to the Koopman coordinates $\psi_k$. Once the intrinsic coordinates are obtained, linear operator $\mathcal{K}$, known as the Koopman operator, is subsequently applied on $\psi_k$ to achieve $\psi_{k+1}$, the time marched Koopman coordinates. The coordinates are then transformed back to the original state vector of ${k+1}^{th}$ time step through the inverse mapping, $\psi^{-1}$. As already stated, we here propose two different deep learning architectures.
\par
\begin{figure}[ht!]
   \centering
    \includegraphics[width=.8\textwidth]{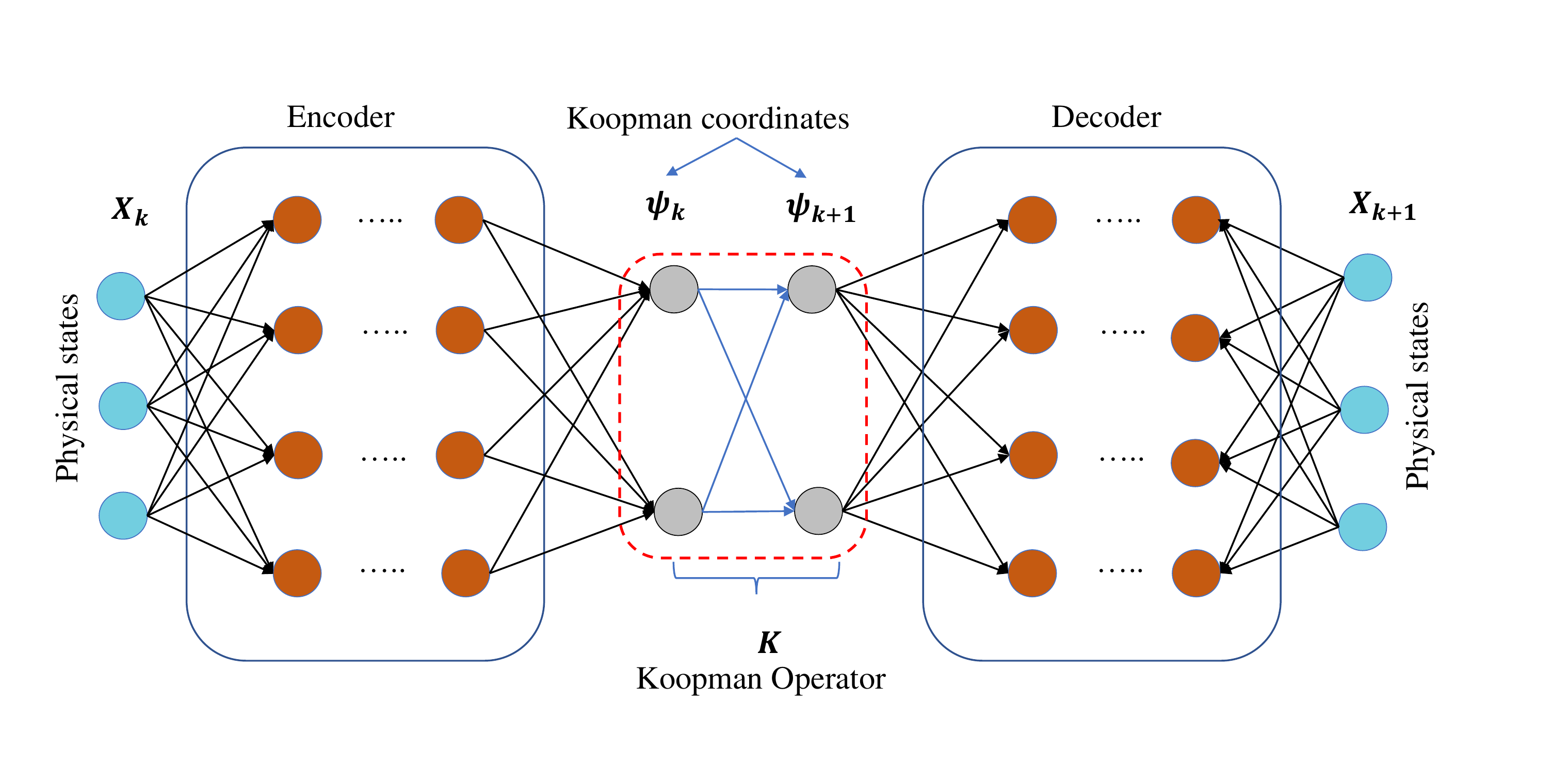}
    \caption{Schematic representation of the proposed architecture for systems with uncertainty in the initial conditions, consists of encoder, decoder and Koopman operator. Training of encoder, Koopman operator and decoder are done simultaneously.}
    \label{fig:Koopman_general1}
\end{figure}
\begin{figure}[ht!]
   \centering
    \includegraphics[width=.8\textwidth]{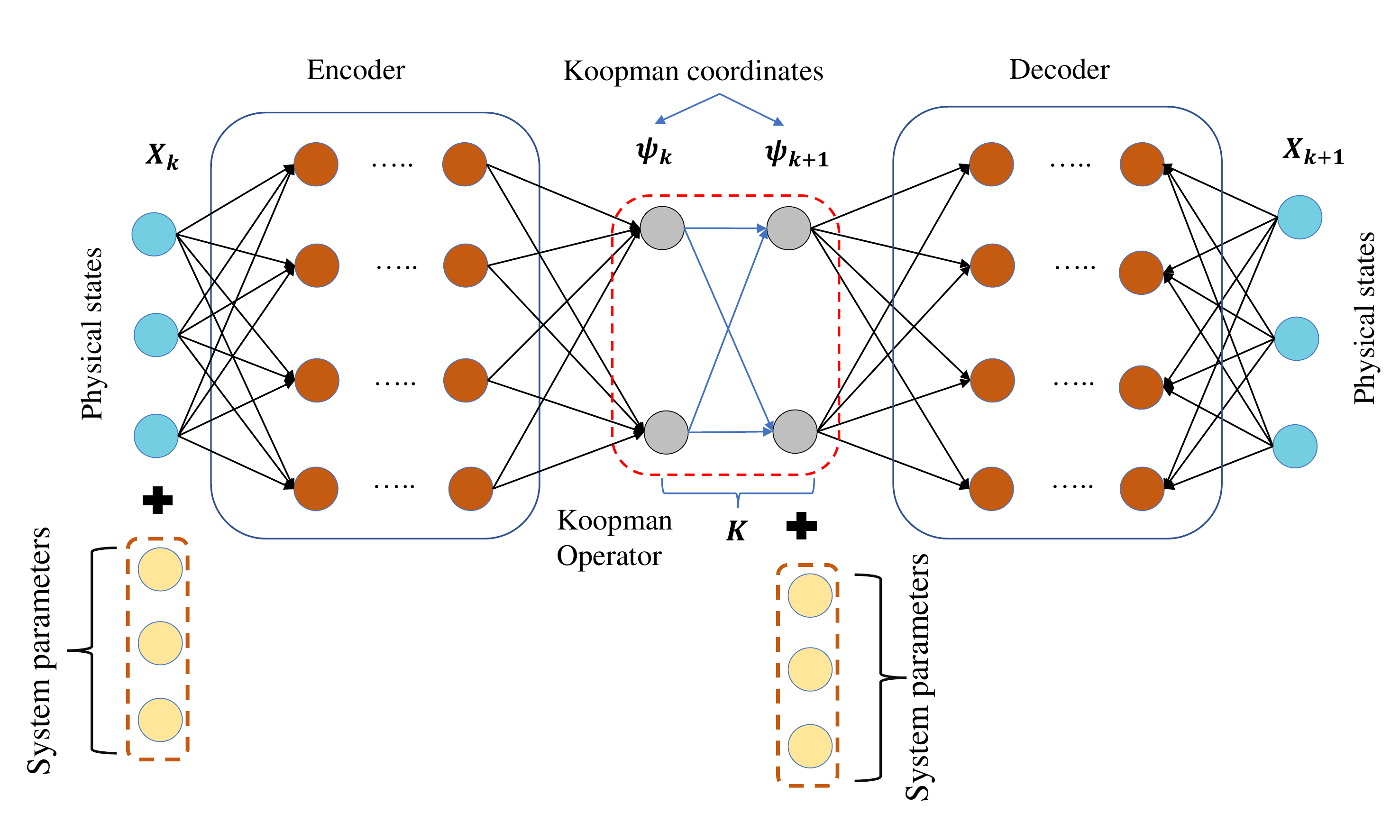}
    \caption{Schematic representation of the proposed architecture for systems with uncertainty in the parameters, consists of encoder, decoder and Koopman operator. State variables appended to system parameters is fed as input to the encoder. Input to the decoder is concatenated vector of Koopman coordinates of successive time step and system parameters}
    \label{fig:Koopman_general2}
\end{figure}

The schematic depiction of the proposed architecture for dynamical systems with uncertainty in the initial conditions is shown in \autoref{fig:Koopman_general1}. The framework is composed of three main parts: encoder, decoder, and Koopman operator. Function of the  encoder is to enable a mapping from the state vector, $\bm X_k$ to the Koopman coordinates $\psi_k(\bm X)$. Further, a subsequent operation through $\mathcal{K}$ yields the $\psi_{k+1}$, where Koopman operator $\mathcal{K}$ is designed as a linear network in the conjunction between the encoder block and decoder block. Finally, the decoder maps back the intrinsic coordinate vector $\psi_{k+1}$ to system state vector $\bm X_{k+1}$.
The second architecture, the framework to achieve the Koopman dynamics for the systems with uncertainty in the model parameters, is schematically represented in \autoref{fig:Koopman_general2}. The design of the framework is similar to the first architecture. Nonetheless, there is a significant difference in regards to inputs of the encoder and decoder. Instead of providing the state vector $\bm X_k$ alone as the input to the encoder, the resulting vector formed by concatenating parameters ($\bm Y$) with the state vector is passed here. Additionally, the concatenated vector of the Koopman coordinates and the parameters are provided as input to the decoder. 
%-------------------------------------------
\par
To train the neural network, an appropriate loss function needs to be defined. The overarching goals of the network architecture includes identifying Koopman coordinates, Koopman operator and future state prediction. As discussed above, while the encoder is designed to identify latent variable as the intrinsic coordinates $\bm Z = \psi(\bm X)$ corresponding to the physical states $\bm X_k$, the decoder seeks a inverse mapping  $\bm X = \psi^{-1}(\bm Z)$ to retrieve the actual state variables from intrinsic coordinates. Thus, the first components of the loss function can be defined as:
\begin{equation}
    \mathcal{L}_1=\|\bm X-\psi^{-1}(\psi(\bm X))\|
\end{equation}
This component of the loss is known as the reconstruction loss as this ensures the minimum loss is incurred during the to and fro transformation of state coordinates and the Koopman coordinates. Moving forward, the linear prediction is achieved through the operator, $\mathcal{K}$ acting on the Koopman coordinates, .i.e.,  $\psi_{k+1}=\mathbf \mathcal{K} \psi_{k}$. Thus, a constraint is imposed in between the latent variable to ensure the linear dynamics in the intrinsic coordinates. Therefore the loss function component can be expressed as:
\begin{equation}
    \mathcal{L}_2=\|\psi (\bm X_{k+1})-\mathcal{K}(\psi(\bm X_k))\|.
\end{equation}  
The component of loss function guides the linear layer to learn the Koopman operator. Finally, it is also important to ensure the successive future predictions state are obtained through Koopman dynamics and hence the third component of loss function is defined as
\begin{equation}
    \mathcal{L}_3=\|\bm X_{k+1}-\psi^{-1}(\mathcal{K}(\psi(\bm X_k)))\|.
\end{equation}  
The total loss function imposed to train the network can be expressed as the sum of the three components of losses,
\begin{equation}\label{eq1:loss}
    \mathcal{L}=\lambda_1 \mathcal{L}_1 + \lambda_2 \mathcal{L}_2 + \lambda_3 \mathcal{L}_3
\end{equation}
Once the loss functions are defined, optimum values of the network parameters are obtained from the minimization of the total loss $\mathcal{L}$. Here, multipliers of each component of the loss function, i.e., $\lambda_1$, $\lambda_2$, and $\lambda_3$ are the hyperparameters of the model and thus they are chosen based on the validation data. Also, it should be noted that, the operator, $\|\cdot\|$ in the expressions of loss functions represents the mean-squared error loss.

Once the network is trained, we use it as a surrogate model for the underlying dynamical system and conduct reliability analysis employing MCS on the trained model. {An algorithm depicting the steps involved are shown in Algorithm \ref{alg:Koopman based reliabilty}}.
\begin{algorithm}
\caption{Koopman operator based time-dependent reliability analysis}\label{alg:Koopman based reliabilty}
{\textbf{Requirements:} Training data set $\{\bm X_{0:t},\bm X_{1:t+1}\}$, multipliers of loss function $\lambda_1$, $\lambda_2$, and $\lambda_3$}, and statistical distributions of the stochastic variables.\\
{\textbf{Output:} PDF of first passage failure time}
\begin{algorithmic}[1]
\State {\textbf{Initialize:} Network parameters $\{w_{i}s, b_{i}s\}$}
\While {{$\mathcal{L} > \epsilon$}}
\State {Train the network:} $\{w_{i}s, b_{i}s\}\leftarrow \{w_{i}s, b_{i}s\}-\delta \nabla_{\bm w, \bm b}{L}(\bm w, \bm b)$
\State {epoch= epoch $+$ 1}
\EndWhile
\For{$i=1,\ldots,N_s$}
\State{Draw $i-$th sample from the stochastic variables}
\State {Obtain $\hat{X}^i_{1:t+1}$ from the initial conditions $X_{0}$ and realization stochastic variable}
\State{Obtain FTTP}
\EndFor
\State {Obtain PDF/histogram of FTTP}
\end{algorithmic}
\end{algorithm}
\section{Numerical examples}\label{sec: Numerical example}
In this section, we evaluate the efficacy of the framework with three numerical examples. The overarching goal here is to evaluate the probability density function of the first passage failure time. For that, we consider randomness in the initial conditions and randomness in the system parameter. The numerical examples selected involve dynamical systems described by ordinary differential equations (ODEs) and partial differential equations (PDEs). While probability density function plot of first passage failure time depicts the failure probability of system over the service period, total failure probability ($P_f$) shows the first passage crossing rate, where corresponding reliability index ($\beta$) is given by:
\begin{equation}
    \beta=\Phi^{-1}(1-P_f)
\end{equation}

\subsection{Duffing oscillator}
\begin{figure}[!ht]
    \centering
    \includegraphics[width=.6\textwidth]{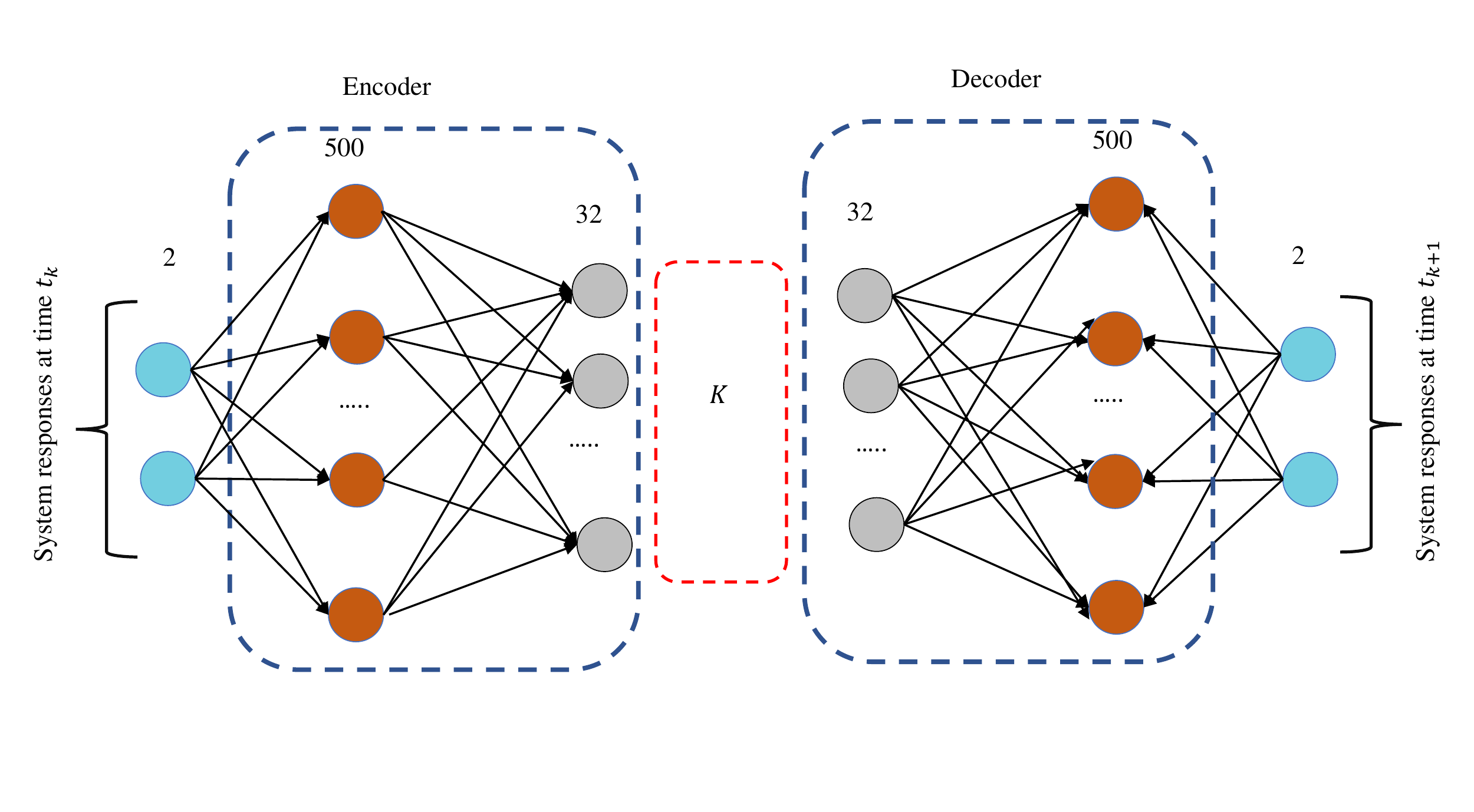}
    \caption{Architecture of proposed Koopman framework (a fully connected network) for duffing oscillator example with  uncertainty in the initial conditions}
    \label{fig:nnarch1}
\end{figure}

As the first numerical example, we illustrate the proposed framework for reliability analysis of the duffing oscillator, which is a benchmark nonlinear system that has drawn the attention of researchers due to its unique dynamical behaviour. The system is described by the differential equation;
\begin{equation}
    \ddot x + \delta \dot x + \alpha x + \beta {x^3} = \gamma \cos (\omega t){\mkern 1mu}
\end{equation}
For the given system we employ the data-driven Koopman framework to predict the response. Once the response is obtained the first passage failure is evaluated. The efficacy of the model is evaluated for the two test cases; firstly, for random initial conditions with constant system parameters and secondly, for random system parameters with initial conditions kept invariable. For the first test case, random initial state are chosen as $x \sim{\mathcal{U}(-5,5)}$ and $\dot x \sim{\mathcal{U}(0,10)}$. 
The architecture for the proposed model is shown in \autoref{fig:nnarch1}. The schematic represents the fully connected auto-encoder model with Koopman coordinates of 32 dimensions. While the encoder block maps the current physical state of dimension 2 to 32, the decoder block transforms back the Koopman coordinates to physical states of dimension 2. Here, the ReLU activation function is applied on hidden layers. The additional linear layer that bridges the encoder-decoder blocks, constitutes the Koopman operator.
\par
\begin{figure}[!ht]
    \centering
    \subfigure[]{
    \includegraphics[width=.4\textwidth]{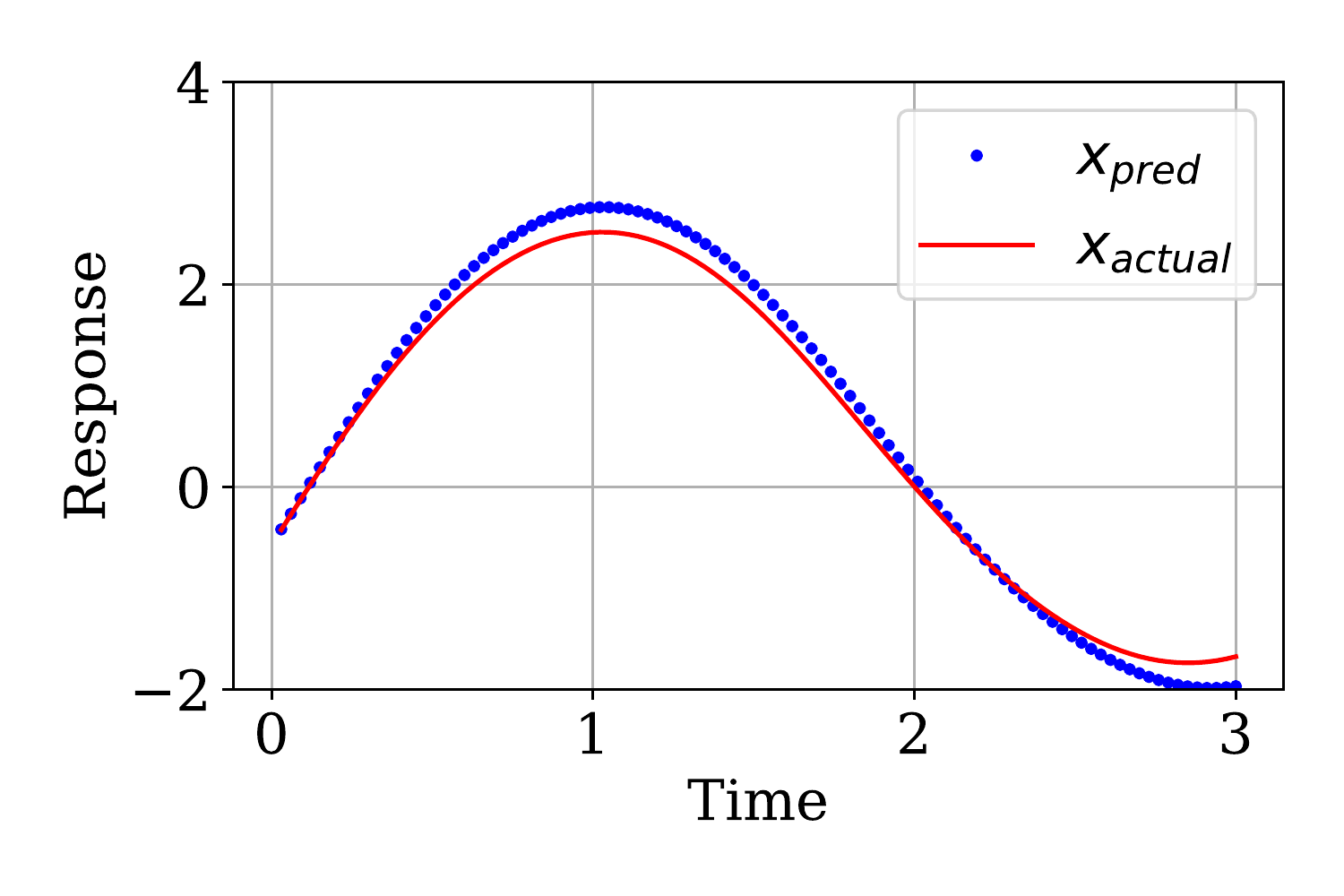}}
    \subfigure[]{
    \includegraphics[width=.4\textwidth]{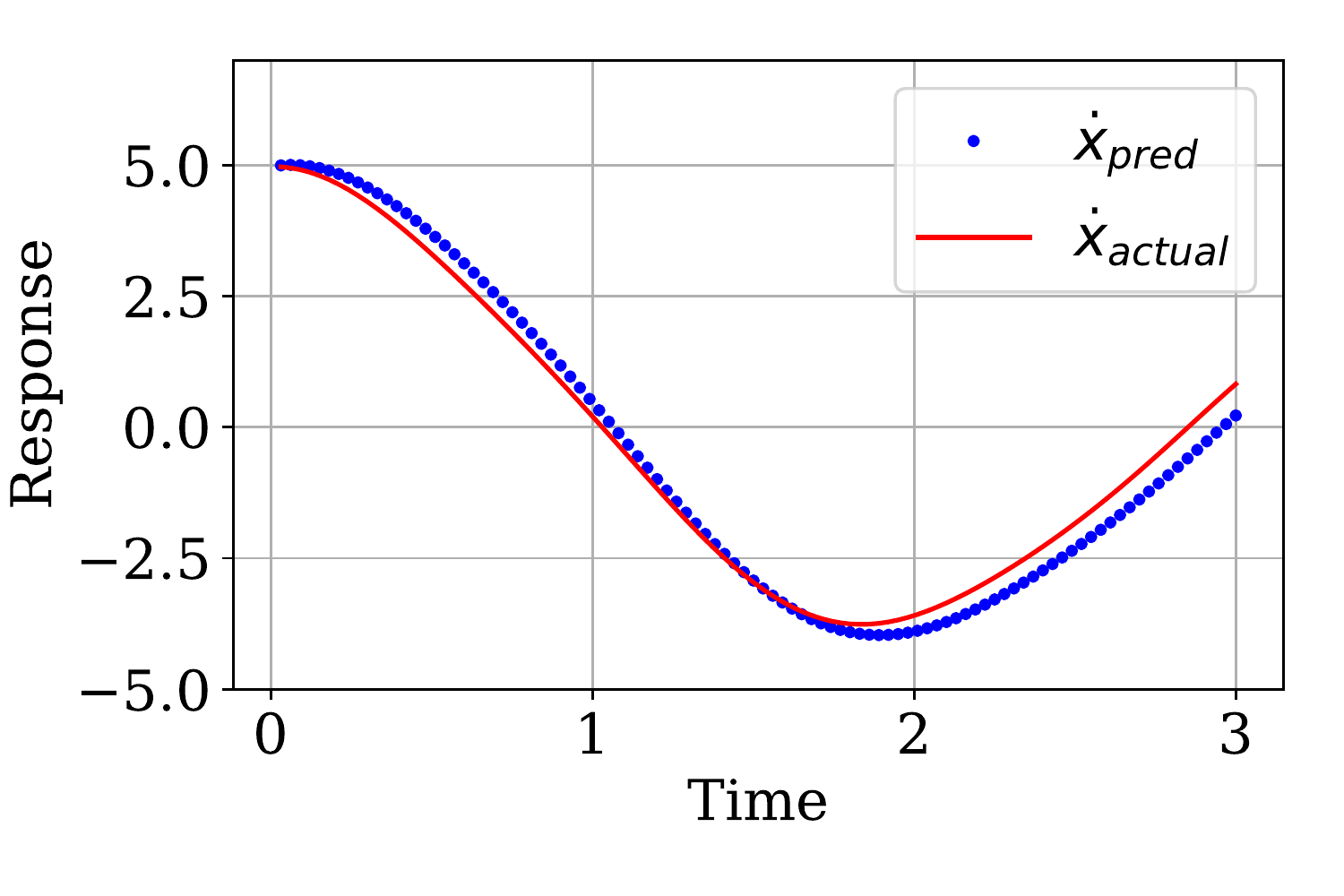}}
    \caption{State variable of duffing oscillator predicted for random initial conditions; (a) variation of displacement ($x$) with time, (b) variation of velocity ($\dot x$) with time }
    \label{fig:case11}
\end{figure}

The framework is provided with training and validation data set of $1800$ and $200$ time series containing 100-time steps. The model is trained for 250 epochs and a learning rate of $10^{-5}$ is used. For the prediction of the time series response, we only provide the initial conditions of the system. The results obtained are shown in the \autoref{fig:case11}. It should be noted that the mean response of predicted data and validation data are presented in the plot. From \autoref{fig:case11} it is evident that the predicted responses are in good agreement with actual values.

\begin{figure}[!ht]
    \centering{
    \includegraphics[width=.4\textwidth]{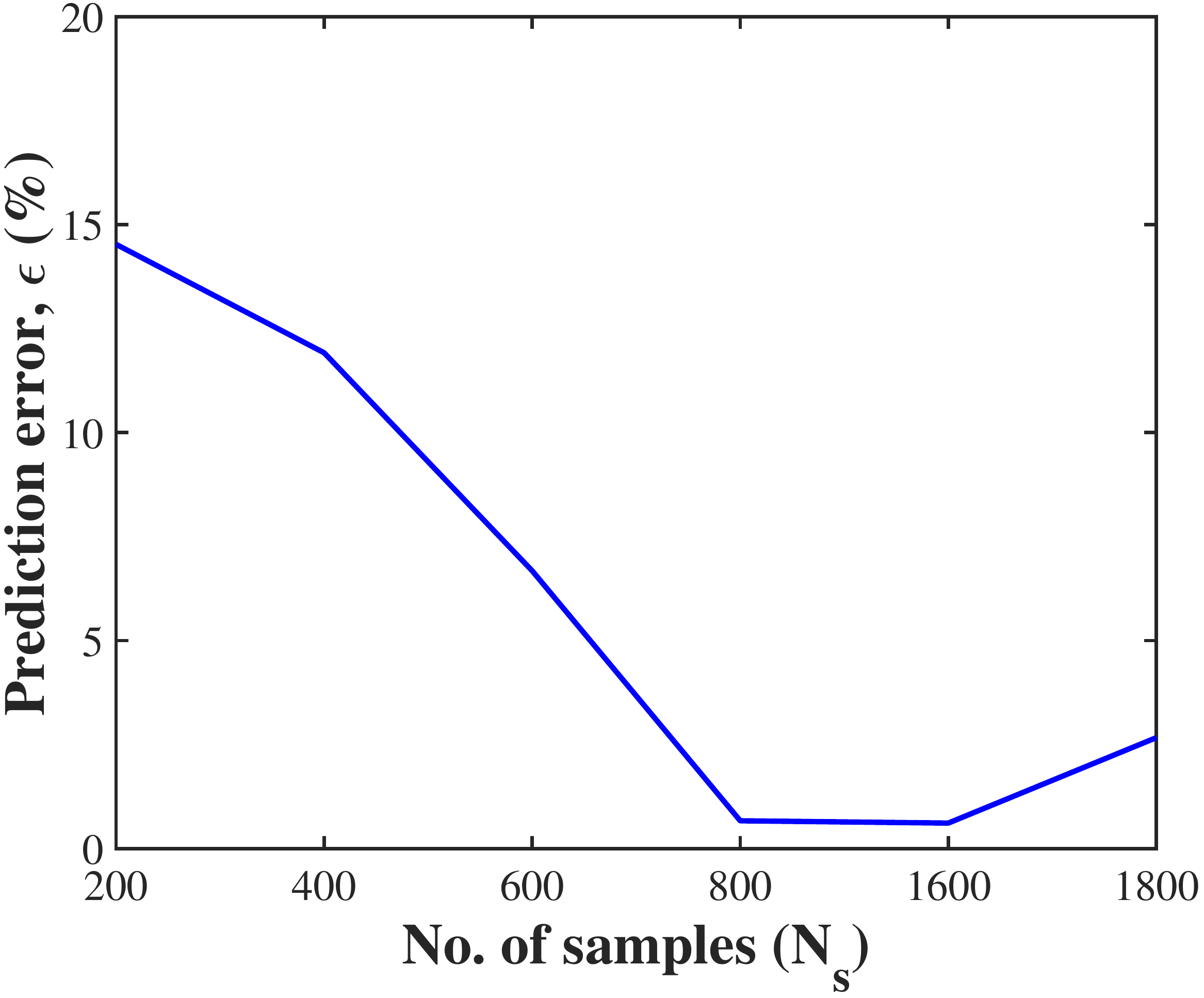}}
    \caption{Convergence of prediction error ($\epsilon $) with number of training samples $N_s$ for the case of duffing oscillator with random initial conditions}
    \label{fig:conv}
\end{figure}

The number of training samples for the proposed framework are chosen based on the convergence study as shown in \autoref{fig:conv}. We note that only 800 training samples are sufficient when the distribution of the random variable remains same for training and test data. However, for better generalization and to enable out-of-distribution capabilities in the proposed framework, 1800 training samples are needed.

\begin{figure}[!ht]
    \centering
    \subfigure[]{
    \includegraphics[width=.4\textwidth]{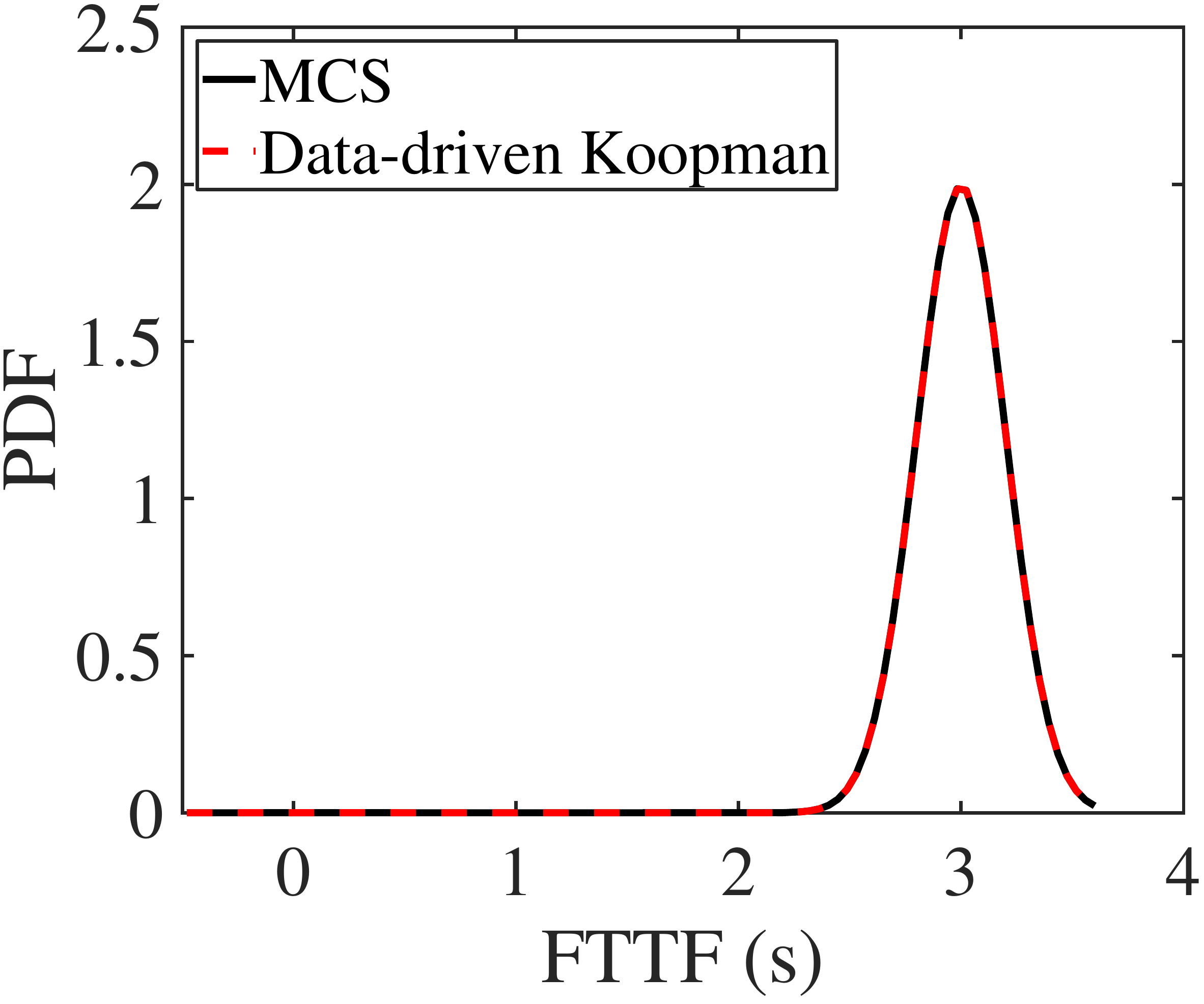}}
    \subfigure[]{
    \includegraphics[width=.4\textwidth]{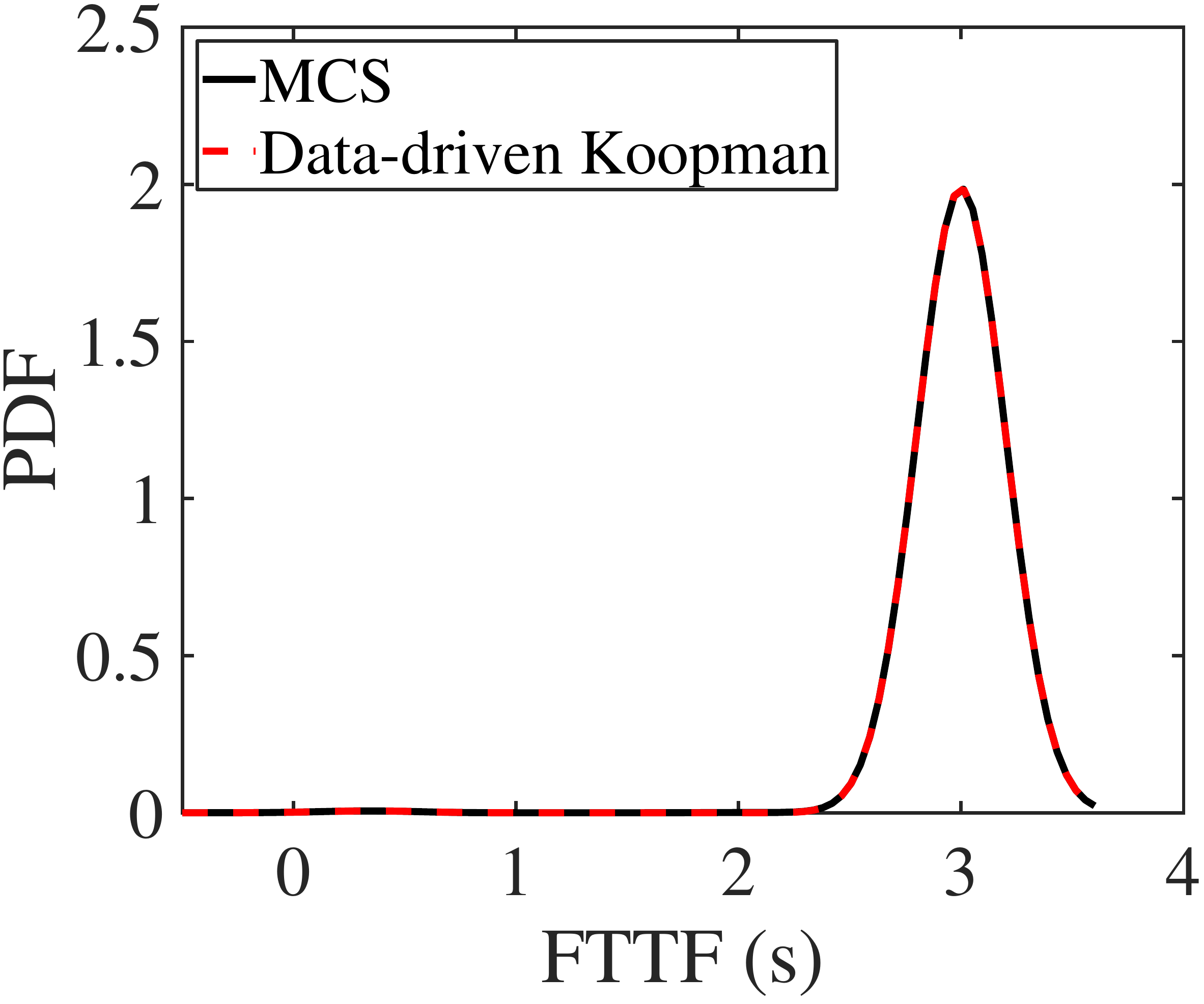}}
    \subfigure[]{
    \includegraphics[width=.4\textwidth]{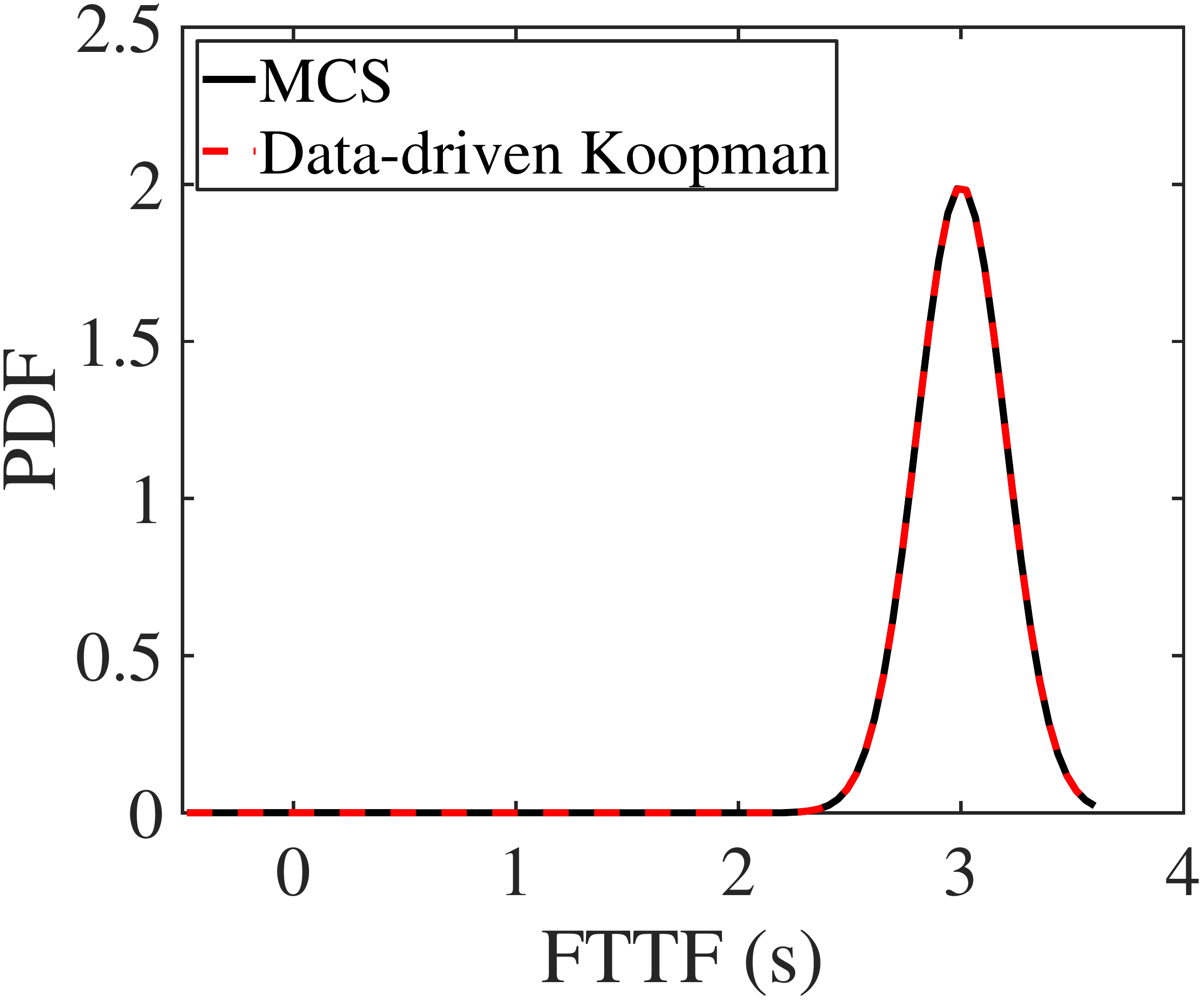}}
    \subfigure[]{
    \includegraphics[width=.4\textwidth]{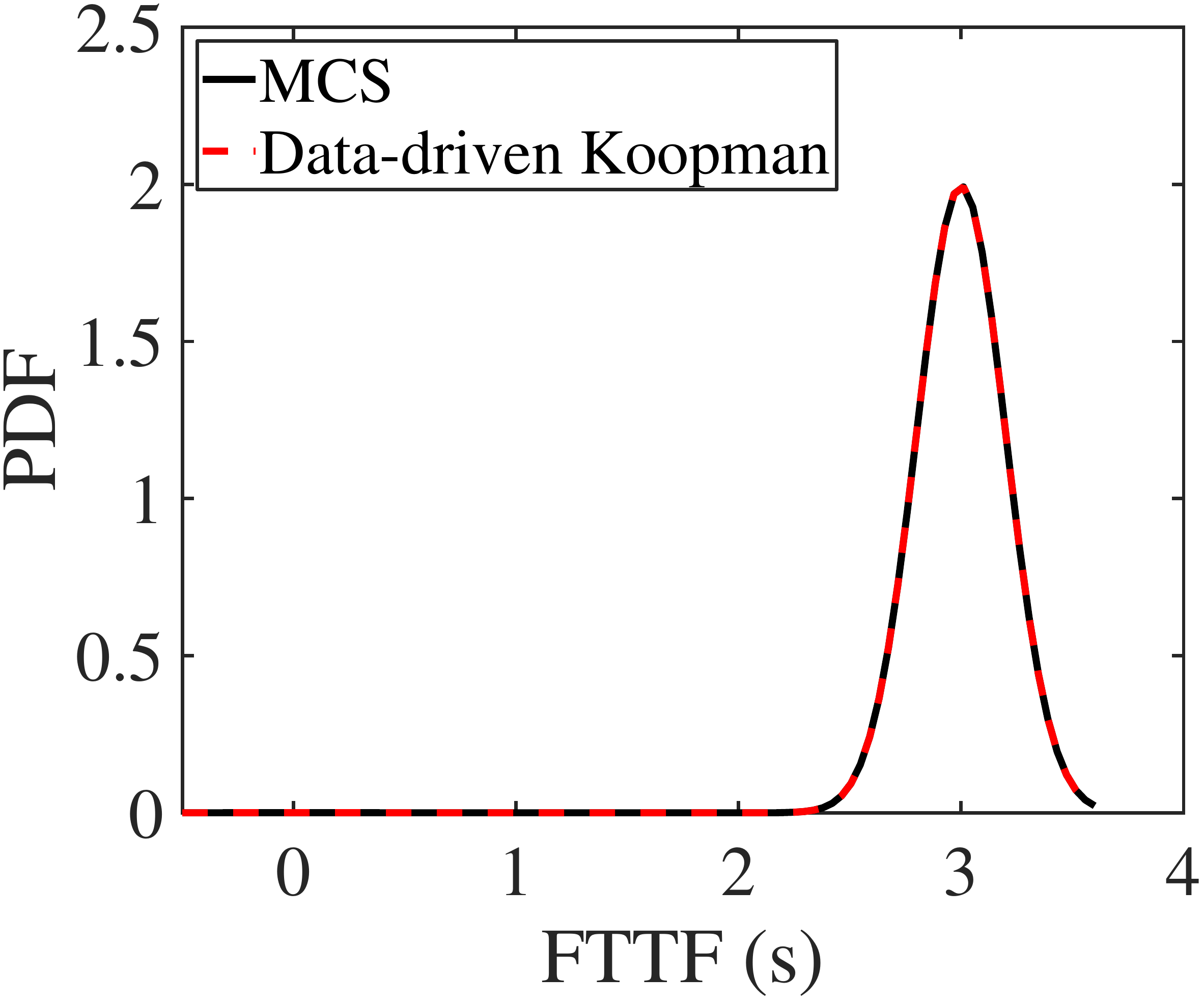}}
    \caption{PDF of failure time obtained by MCS and proposed framework for the case of  uncertainties in the initial conditions, with samples state variables generated in $x\sim(-5,5)$ and $\dot x\sim(0,10)$ having distribution (a)uniform (b) Gaussian (c) lognormal, and with samples generated $x\sim(-6,6)$ and $\dot x\sim(0,12)$ having distribution (d) uniform (e) Gaussian (f) lognormal}
    \label{fig:cased11}
\end{figure}

Fig. \ref{fig:cased11} shows the probability density function (PDF) of the first passage failure time for the random initial condition case obtained using the proposed framework, auto-regressive fully connected neural network (FNN), and Long Short-Term Memory (LSTM). To illustrate the robustness of the proposed approach, different cases have been considered. 
% Here, two sets of test samples consisting of $10^4$ times series having 100 times steps with uniformly distributed initials conditions are  utilized obtain the PDF.
Benchmark results have been generated using Monte-Carlo Simulation (MCS).
The threshold here is $e_h=6$ for all the sets of test samples and accordingly, failure is defined when the displacement response crosses the threshold for the first time. Though the framework can predict the two states of the duffing oscillator $x$ and $\dot{x}$, failure is defined only based on $x$. We observe that the proposed approach yield highly accurate results and matches almost exactly with the benchmark MCS results for all the cases. FNN and  LSTM, on the other hand fails to capture the first mode of the PDF for both the cases.

\begin{figure}[!ht]
    \centering
    \subfigure[]{
    \includegraphics[width=.4\textwidth]{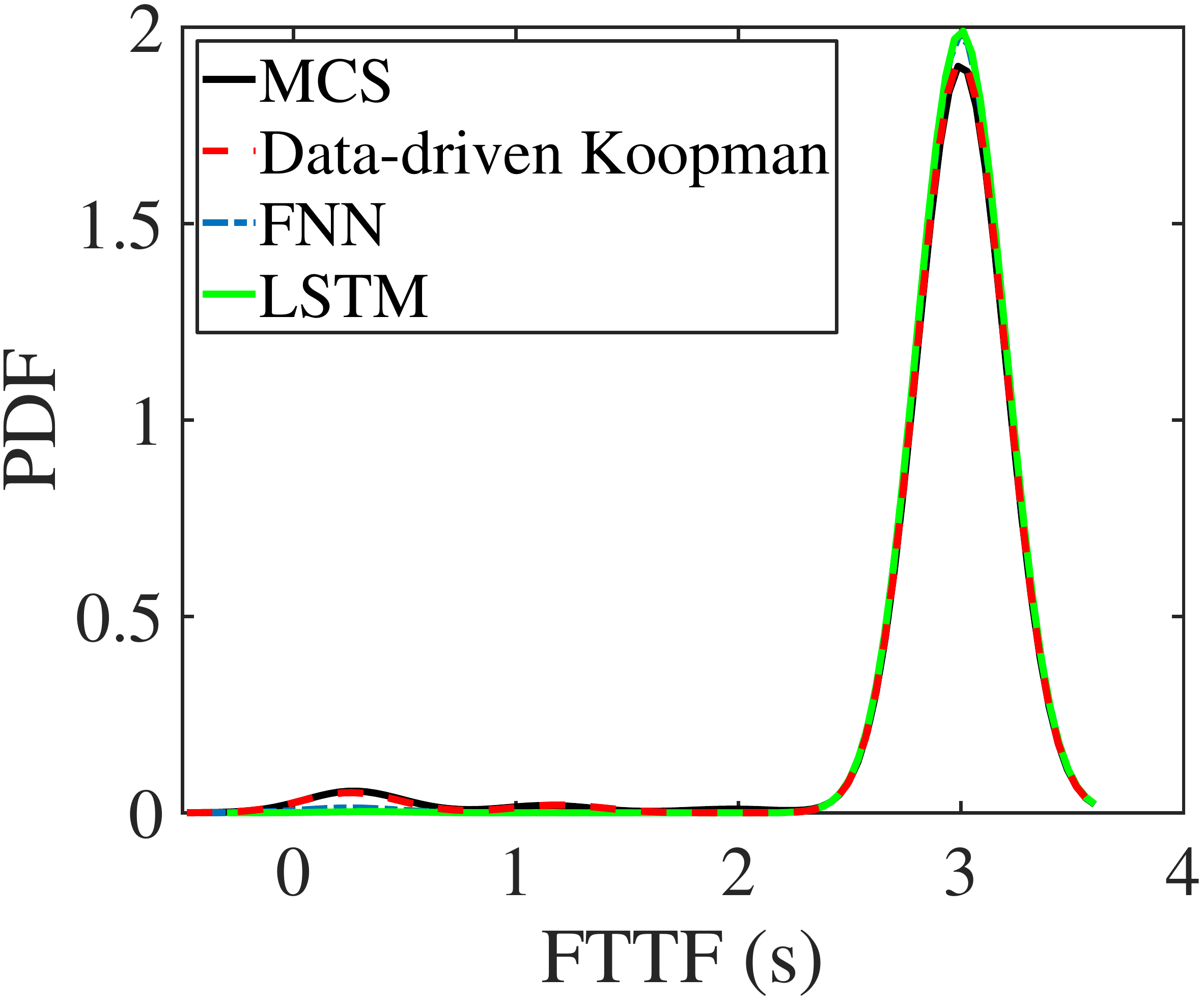}}
    \subfigure[]{
    \includegraphics[width=.4\textwidth]{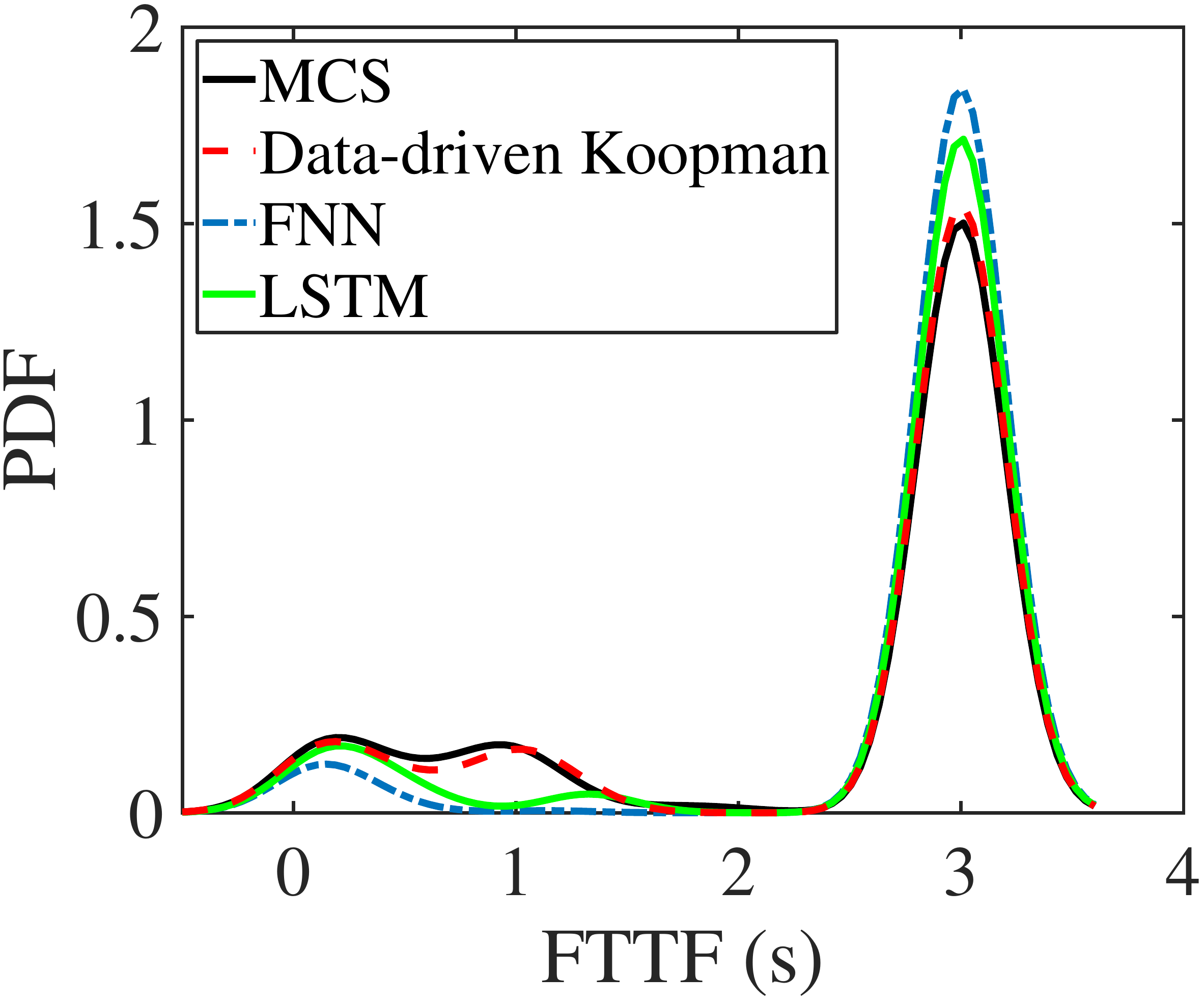}}
    \caption{PDF of failure time obtained by MCS, FNN, LSTM and proposed framework for the case of  uncertainties in the initial conditions, (a) with uniformly distributed samples of state variables generated in $x\sim(-5,5)$ and $\dot x\sim(0,10)$ (b) with samples generated in $x\sim(-6,6)$ and $\dot x\sim(0,12)$}
    \label{fig:cased11e}
\end{figure} 
For further examination, the PDF plots of first passage failure time is also obtained for four different sets of test samples chosen from different distribution (out-of-distribution). While the first two sets of test samples are generated in the support interval of $x\sim(-5,5)$ and $\dot x\sim(0,10)$ having distributions Gaussian and log-normal, the other two sets of test samples are generated in the support interval of $x\sim(-6,6)$ and $\dot x\sim(0,12)$ having distributions Gaussian and log-normal. \autoref{fig:cased11e} reinforces the efficacy the proposed framework in estimating the first passage failure time. 
In addition to probability distribution plots of first passage failure time, total failure probability and corresponding reliability index are obtained by fixing the threshold, $e_h=6$ and considering the first 40 time steps.
\begin{table}[ht!]
    \centering
    \caption{Results of first passage failure probability obtained using    
    proposed-framework for the case of duffing oscillator with random initial conditions. $\beta_e$ indicates the reliability index obtained using MCS.}
    \label{table1}
\begin{tabular}{lcccc} 
\hline
\textbf{Method }& \textbf{Reliability Index} & \textbf{Failure Probability} & \textbf{$N_{s}$}&$\epsilon=\frac{|\beta_{e}-\beta|}{\beta_{e}}\times100$\\
\hline
MCS           &  1.764  &   0.0389 &   $10^{4}$ &-\\ \hline
Koopman         &  1.810 &   0.0351  &  1800 & 2.67\\ 
\hline 
\end{tabular}
\end{table}
\autoref{table1} showcases the results of failure probability and reliability index. The proposed method achieves excellent results in predicting the failure probability with a prediction error of $2.67\%$.
\par
\begin{figure}[!ht]
    \centering
    \includegraphics[width=.6\textwidth]{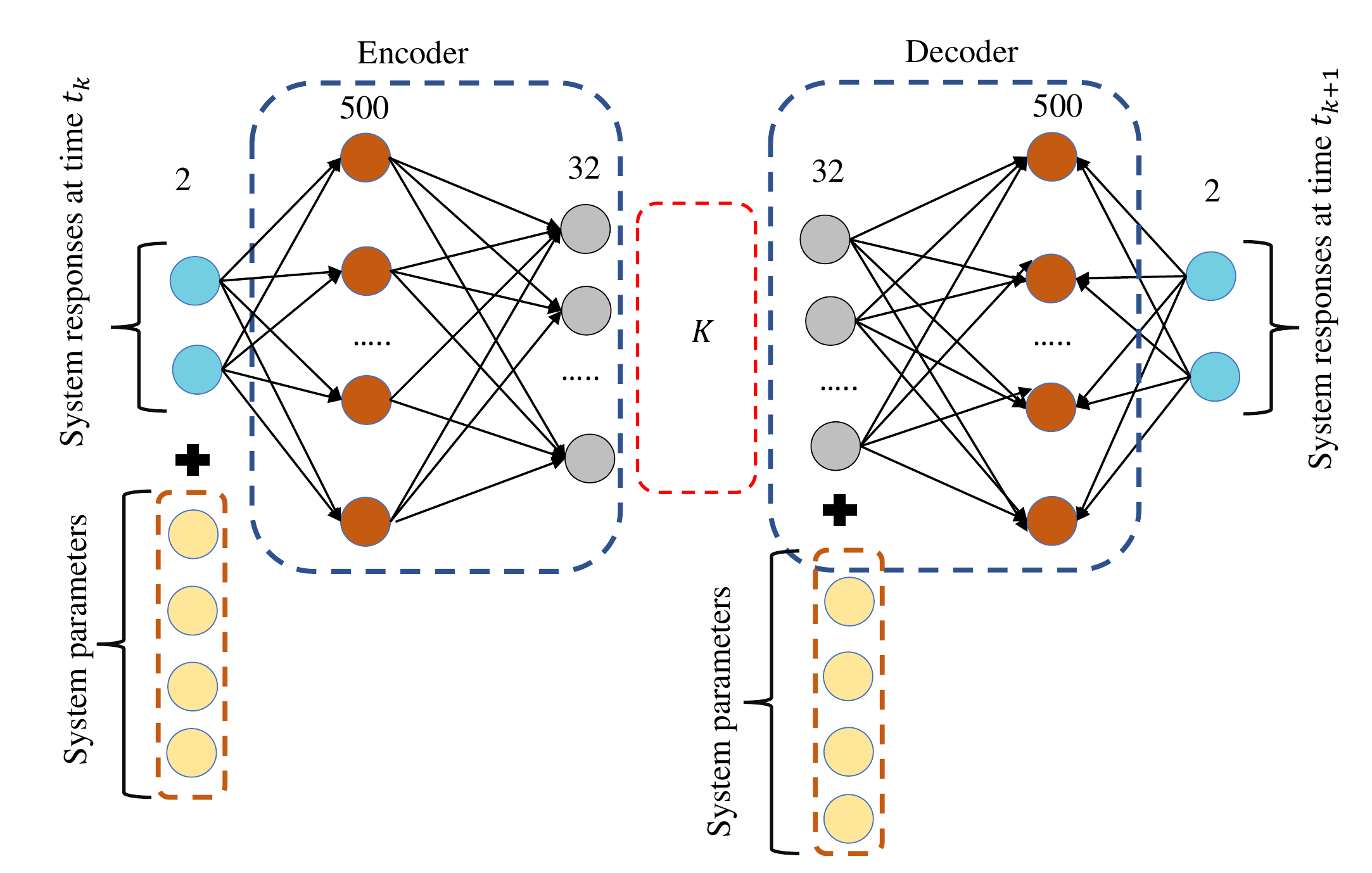}
    \caption{Architecture of proposed Koopman framework for duffing oscillator example with  uncertainty in the system parameters}
    \label{fig:nnarch2}
\end{figure}
\begin{figure}[!ht]
    \centering
    \subfigure[]{
    \includegraphics[width=.4\textwidth]{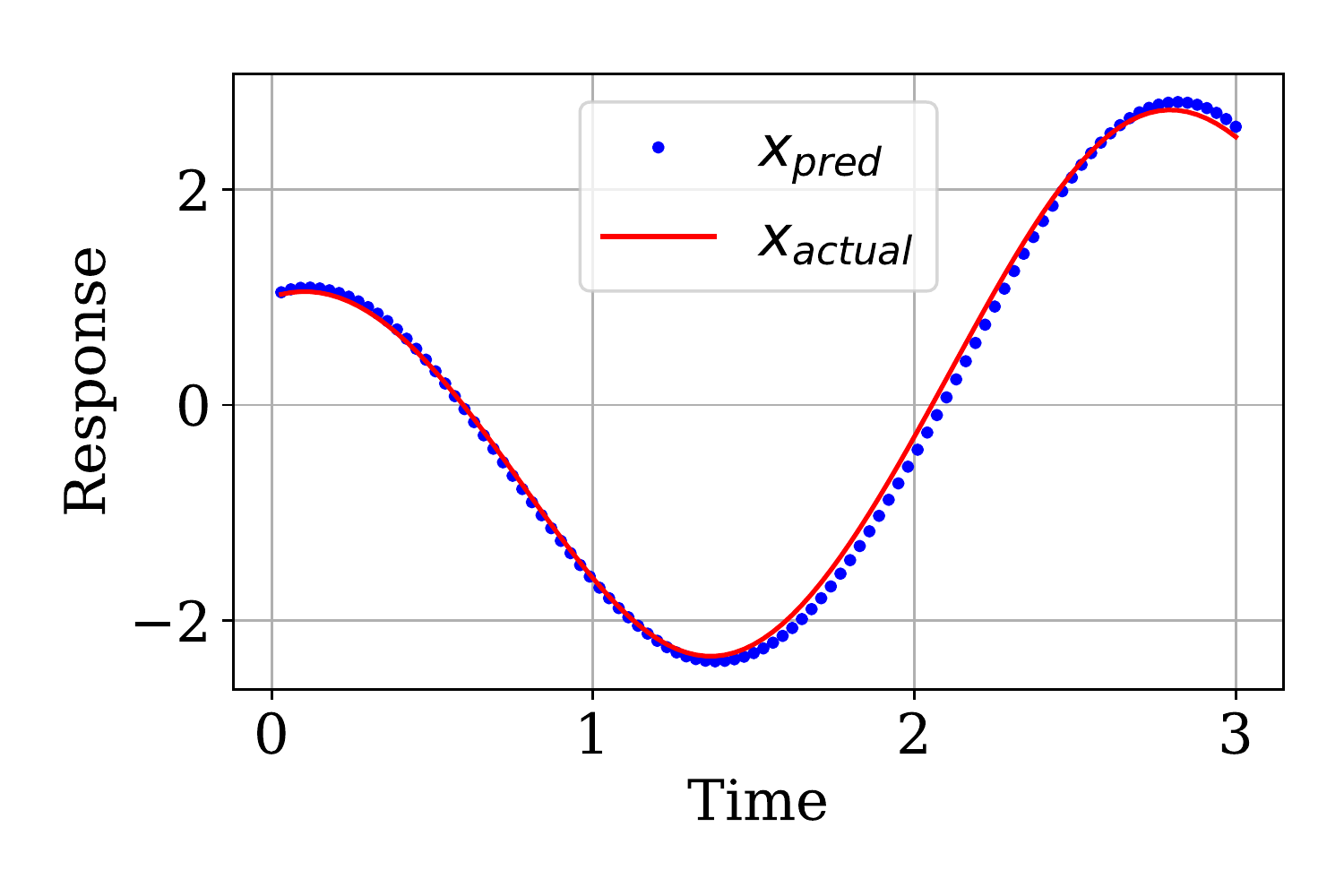}}
    \subfigure[]{
    \includegraphics[width=.4\textwidth]{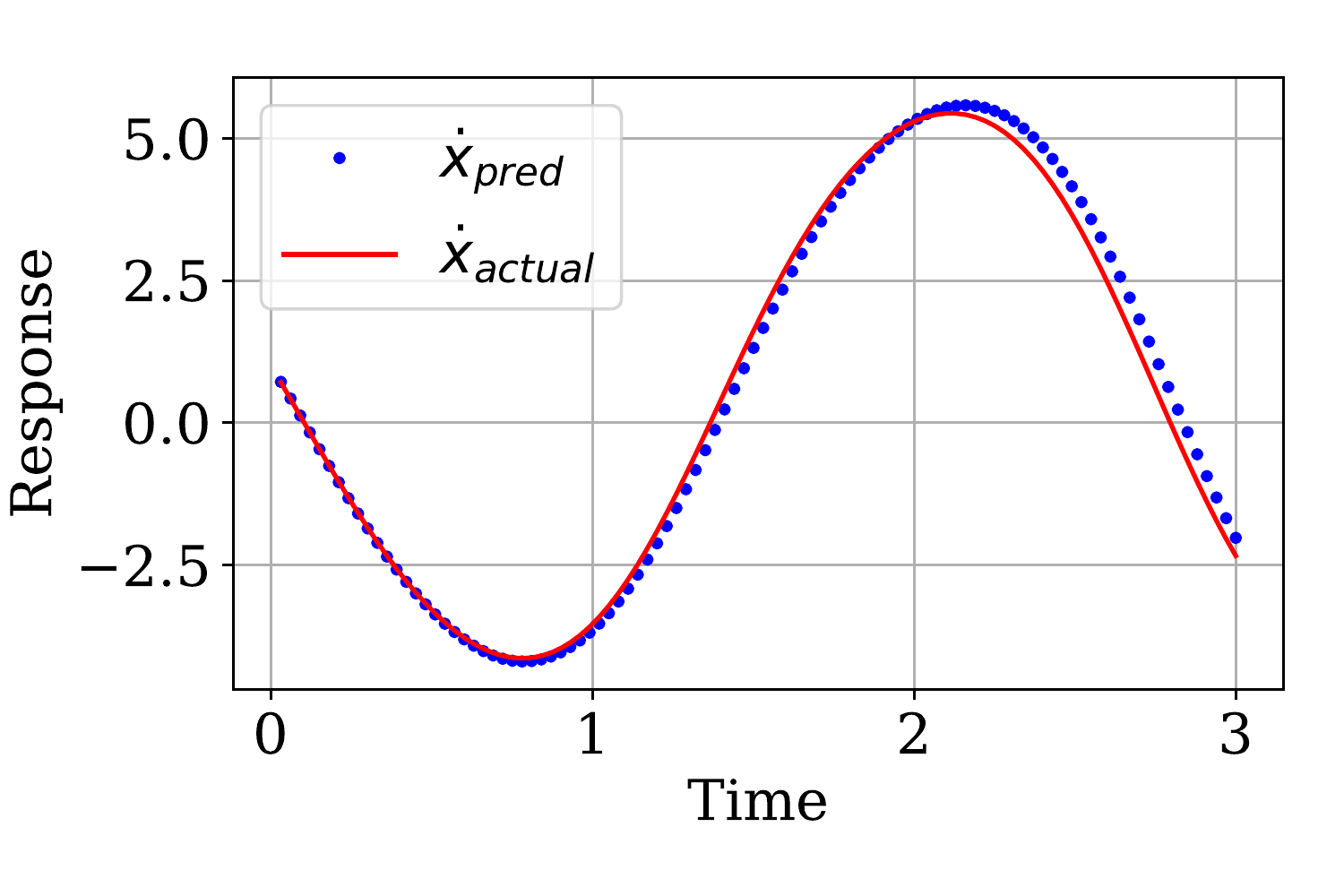}}
    \caption{State variables of duffing oscillator predicted for random system parameters; (a) variation of displacement ($x$) with time, (b) variation of velocity ($\dot x$) with time}
    \label{fig:case12}
\end{figure}

For the second test case, random parameters of the system are varied in the range of $\delta \sim{\mathcal{U}(0.02,0.04)}$, $\alpha \sim{\mathcal{U}(2,6)}$, $\beta \sim{\mathcal{U}(0.1,0.3)}$ and $\gamma \sim{\mathcal{U}(2,8)}$. A training data of $1800$ time series and a validation data of $200$ time series are utilised here. The architecture of the framework for the second case is also an auto-encoder with a ReLU activation and having Koopman coordinates of 32 dimensions. Nevertheless, it is slightly modified to accommodate the system parameters as discussed in \autoref{Koopman_rep}. As the \autoref{fig:nnarch2} shows, the input layer of encoder and decoder consists of additional four neurons corresponding to the four system parameters.
The prediction of responses by the proposed approach is illustrated in \autoref{fig:case12}. The close agreement of the prediction with actual response data is seen in this case as well.

\begin{figure}[!ht]
    \centering
    \subfigure[]{
    \includegraphics[width=.4\textwidth]{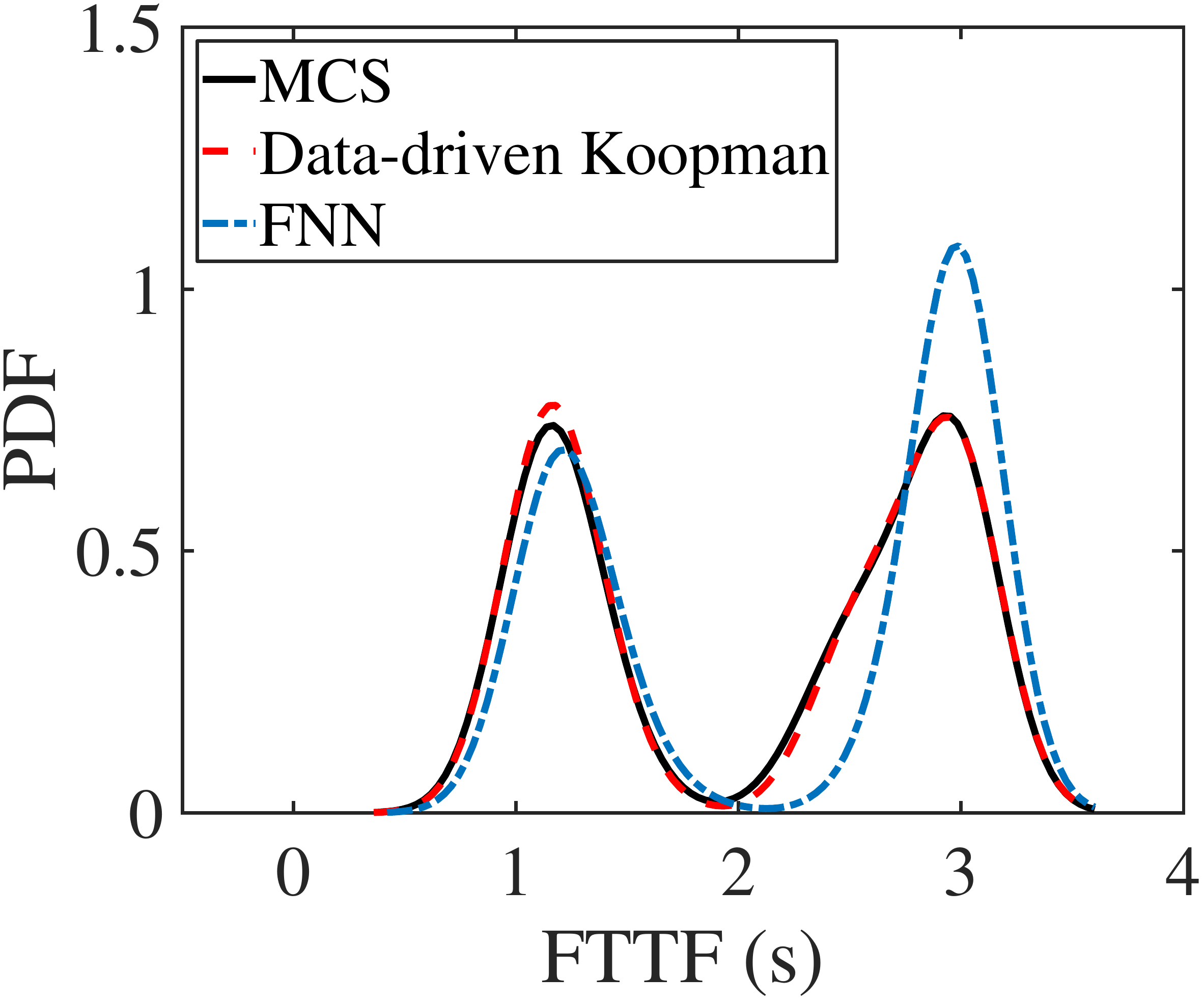}}
    \subfigure[]{
    \includegraphics[width=.4\textwidth]{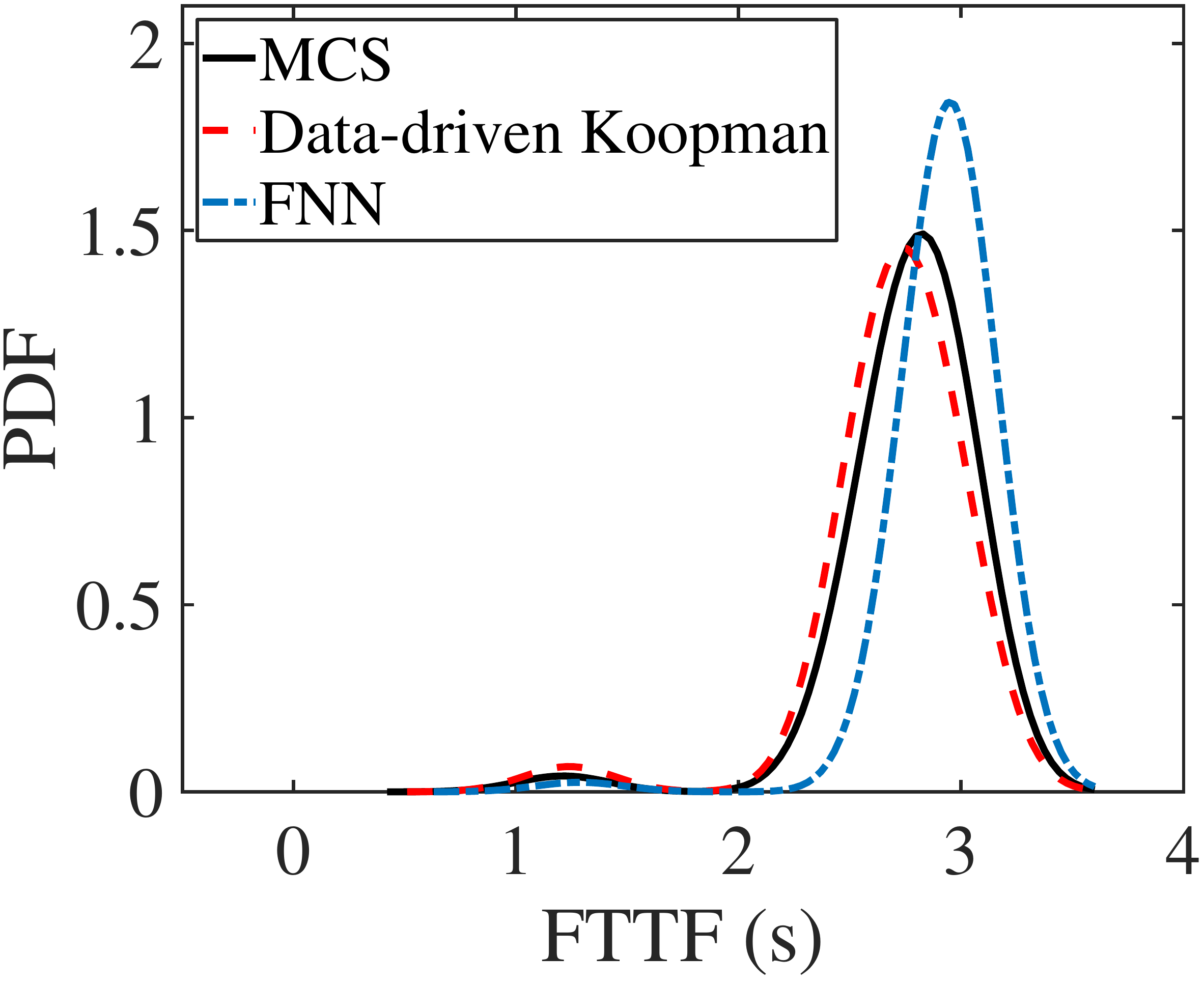}}
    \subfigure[]{
    \includegraphics[width=.4\textwidth]{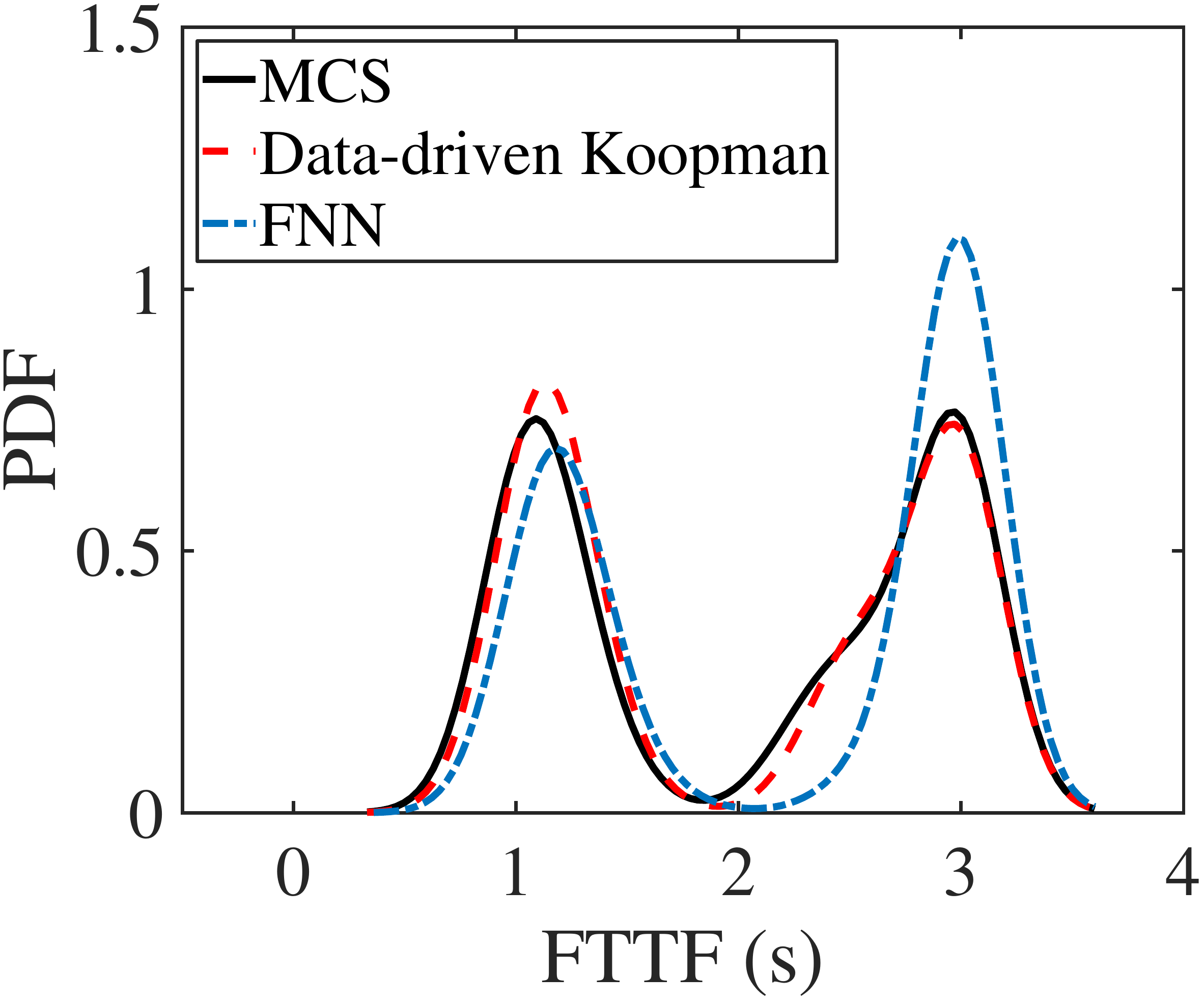}}
    \subfigure[]{
    \includegraphics[width=.4\textwidth]{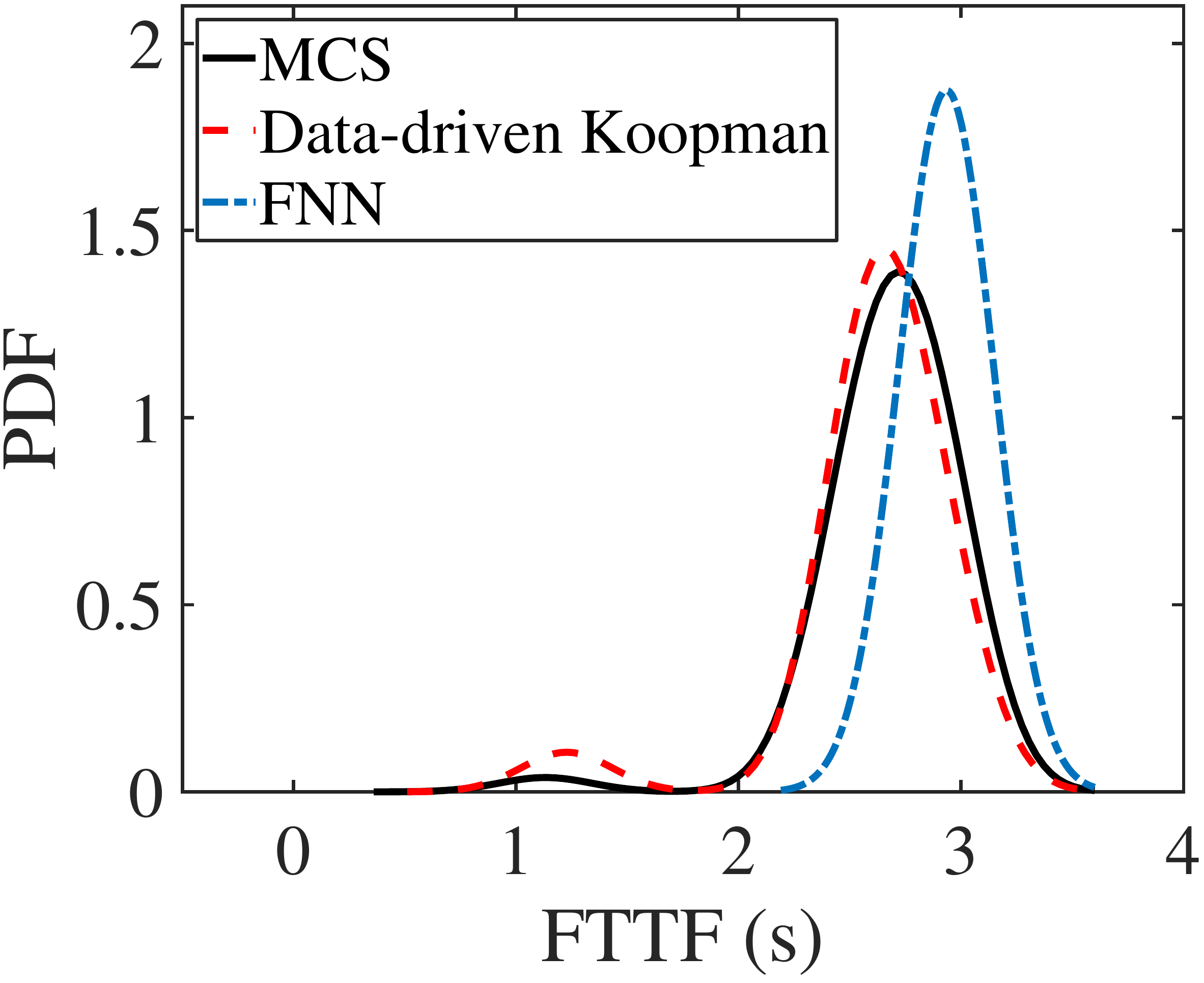}}
    \caption{PDF of failure time obtained by MCS and proposed framework for the case of  uncertainties in the system parameters, with samples of  parameters generated in $\delta  \sim(0.02,0.06),\alpha \sim(2,6),\beta \sim(0.1,0.4)$ and $\gamma \sim(2,8)$ having distribution (a)uniform (b) gaussian (c) lognormal, and with samples generated $\delta \sim(0.02,0.07),\alpha \sim(2,7),\beta \sim(0.1,0.5) $ and $ \gamma \sim(2,9)$ having distribution (d) uniform (e) gaussian (f) lognormal}
    \label{fig:cased12}
\end{figure}

\autoref{fig:case12} shows the PDF plots of first passage failure time in four different sets of test samples of system parameters variations. In regards to the test sets, the number of samples in the test set and length of the time series is chosen the same as that of the previous case. To illustrate robustness of the proposed approach, the range and distributions of the random variables are varied. For all the cases, the threshold is set at $e_h=2.5$. The results of first passage failure estimated by the proposed method are demonstrated in \autoref{fig:cased12}. It can be observed from the plots that, in all four cases, the results obtained by the proposed framework outperforms auto-regressive FNN and is in close agreement with that of MCS.

Fixing the threshold $e_h=2.5$, the total failure probability and accompanying reliability index as given in \autoref{table2}. A reasonable match between the results obtained using the proposed framework and MCS result is observed.

\begin{table}[ht!]
    \centering
    \caption{Results of first passage failure probability obtained using proposed-framework for the case of duffing oscillator with random system parameters. $\beta_e$ indicates the reliability index obtained using MCS.}
    \label{table2}
\begin{tabular}{lcccc} 
\hline
\textbf{Method }& \textbf{Reliability Index} & \textbf{Failure Probability} & \textbf{$N_{s}$}&$\epsilon=\frac{|\beta_{e}-\beta|}{\beta_{e}}\times100$\\
\hline
MCS           &  0.540  &   0.295 &   $10^{4}$ &-\\ \hline
Koopman         &  0.505 &   0.307  &  1800 & 6.4\\ 
\hline 
\end{tabular}
\end{table}

\subsection{Lorenz system}
\begin{figure}[!ht]
    \centering
    \includegraphics[width=.6\textwidth]{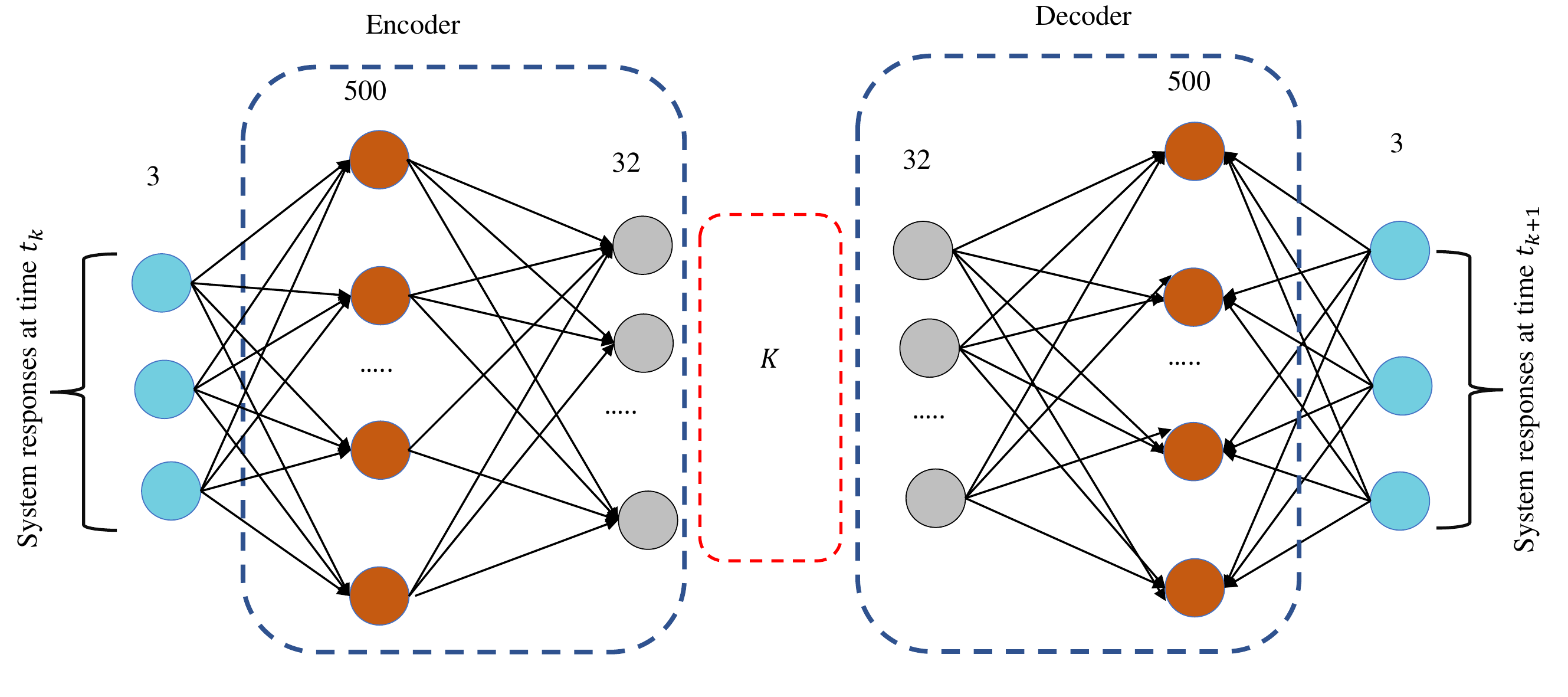}
    \caption{Architecture of proposed Koopman framework for Lorenz system example with  uncertainty in the initial conditions}
    \label{fig:nnarch3}
\end{figure}

\begin{figure}[!ht]
    \centering
    \subfigure[]{
    \includegraphics[width=.4 \textwidth]{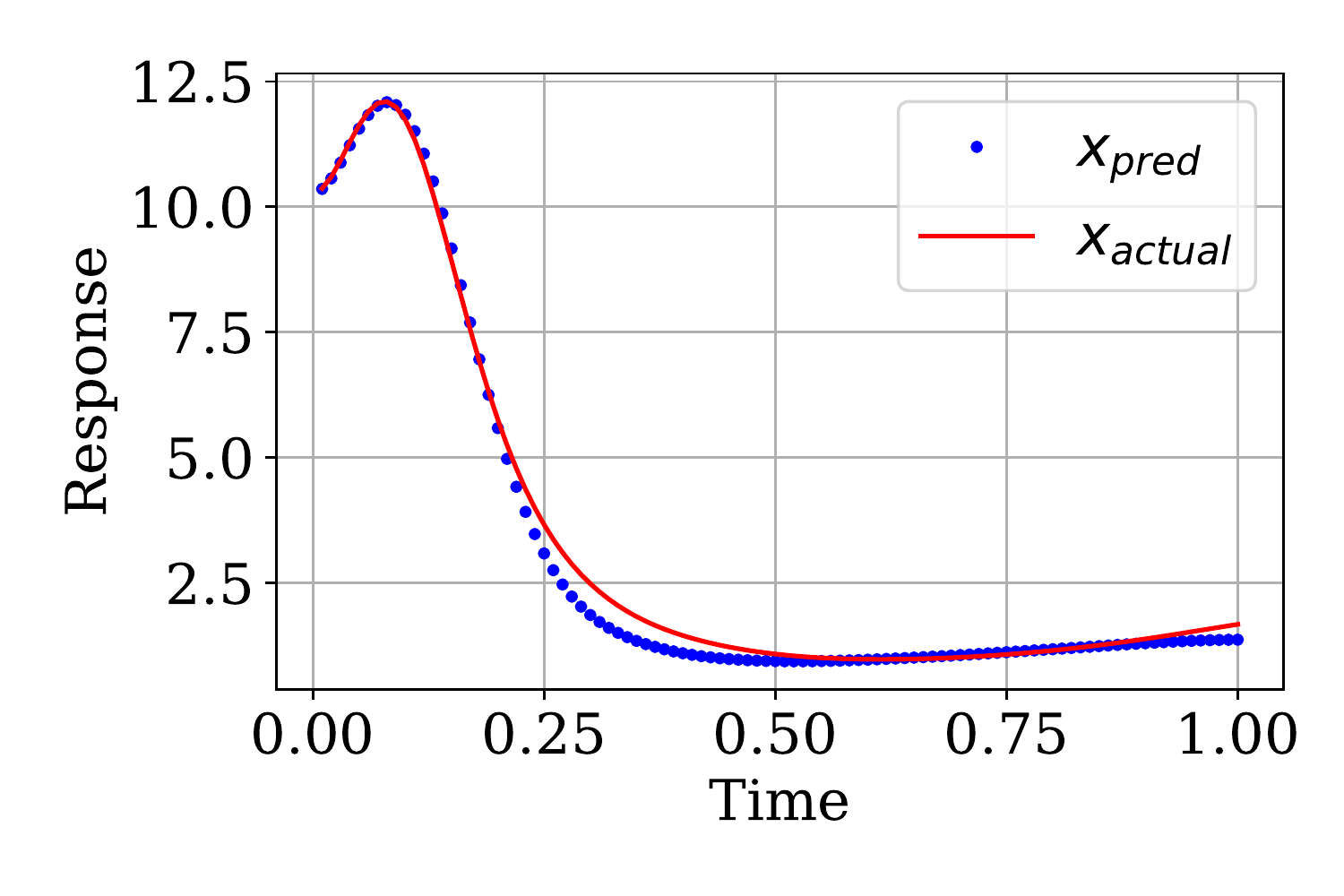}}
    \subfigure[]{
    \includegraphics[width=.4\textwidth]{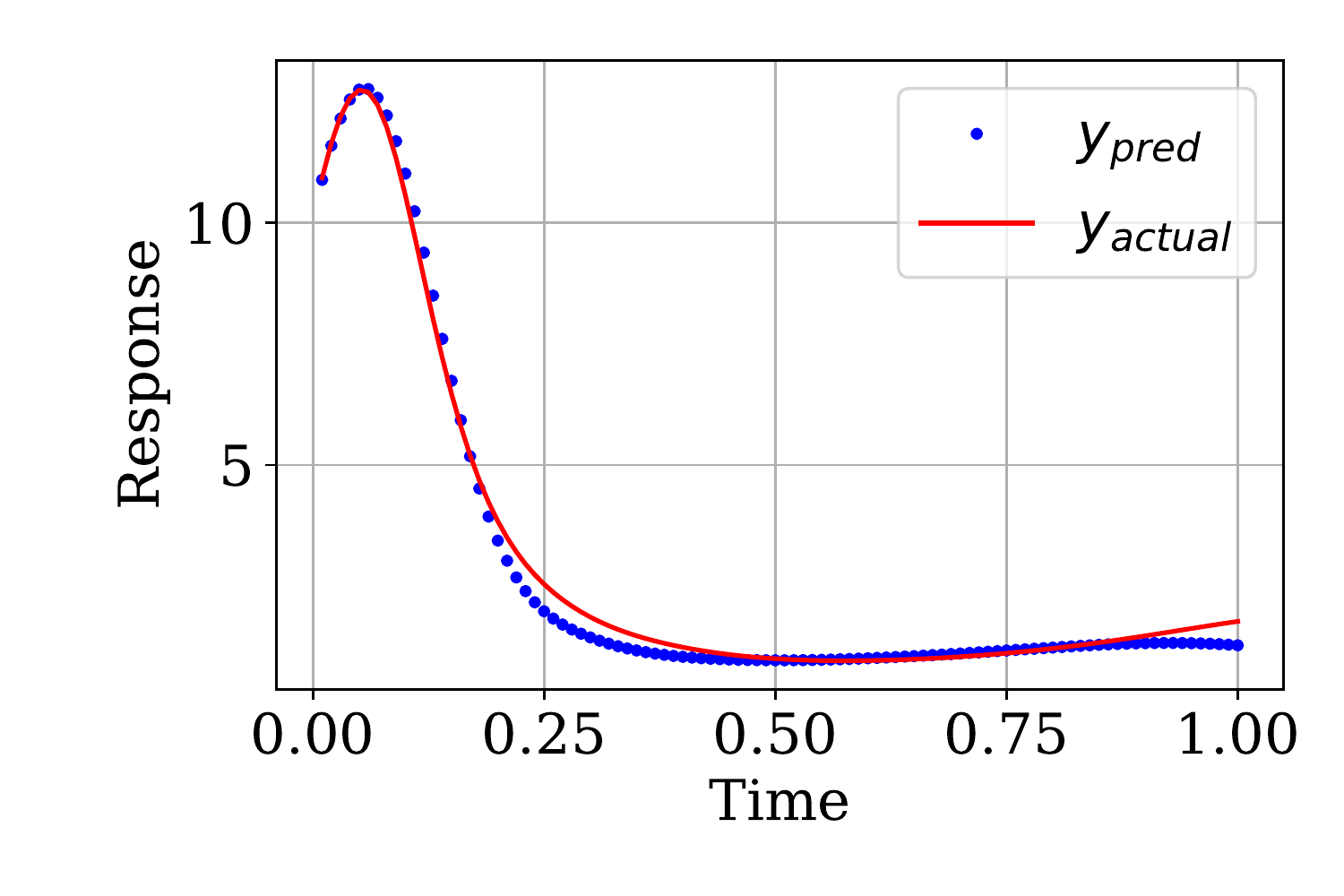}}
    \subfigure[]{
    \includegraphics[width=.4\textwidth]{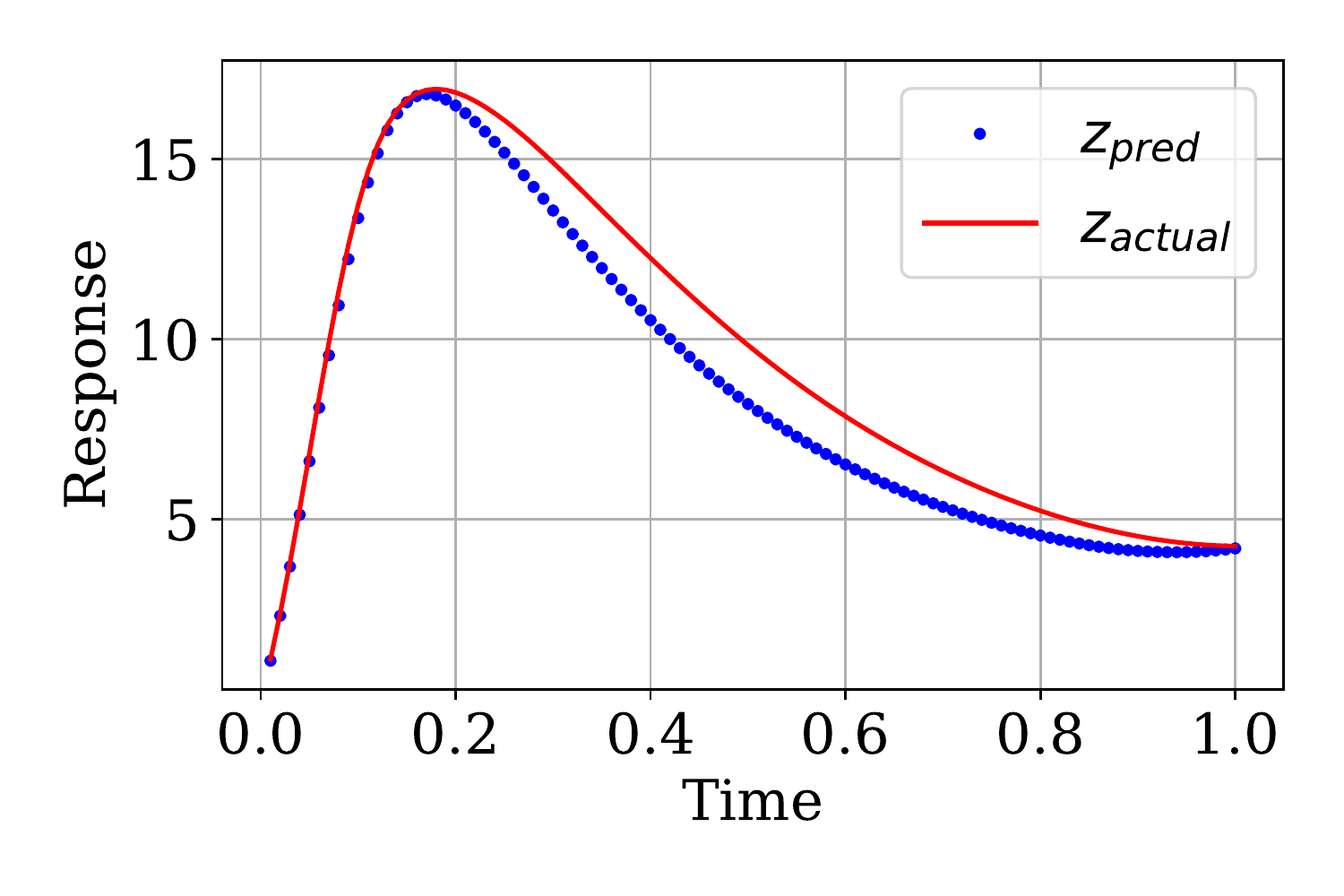}}
    \caption{State variables of Lorenz system predicted for random initial conditions; (a) variation of $x$ with time, (b) variation of $y$ with time, (c) variation of $z$ with time}
    \label{fig:case21}
\end{figure}

The second numerical example we consider is the Lorenz system. This is a chaotic dynamical systems and is highly sensitive to the initial conditions; thus, trajectories of nearby initial states diverge rapidly. The system is expressed in terms of three ordinary differential equations,
\begin{equation}
\begin{array}{l}
\dot x = \sigma(y-x), \\
\dot y = x(\rho-z)-y,\\
\dot z= x y-\beta z.
\end{array}
\end{equation}
Similar to the previous numerical example, the model is evaluated for two cases, random initial conditions as well as random system parameters. Uncertainties in the initial conditions are taken such that $x\sim \mathcal{U}(0,20), y\sim \mathcal{U}(0,20), z\sim \mathcal{U}(-10,10)$.
The proposed framework architecture is depicted in \autoref{fig:nnarch3}. The input layer of the autoencoder has three neurons as the Lorenz system has three state variables and Koopman coordinates are chosen to be of dimension 32. 
While the training set consists of $1800$ time-series validation set consisting of $200$ time series having 100-time steps. Once the model is trained for 150 epochs, initial conditions of the state variable are provided to the framework to predict time series. The results prediction of responses is shown in the \autoref{fig:case21}. While excellent match is observed for the state variables $x$ and $y$, some deviations in $z$ is observed.
%Results obtained are presented in \autoref{fig:case21} and \autoref{fig:case22}. 
\par
\begin{figure}[!ht]
    \centering
    \subfigure[]{
    \includegraphics[width=.4\textwidth]{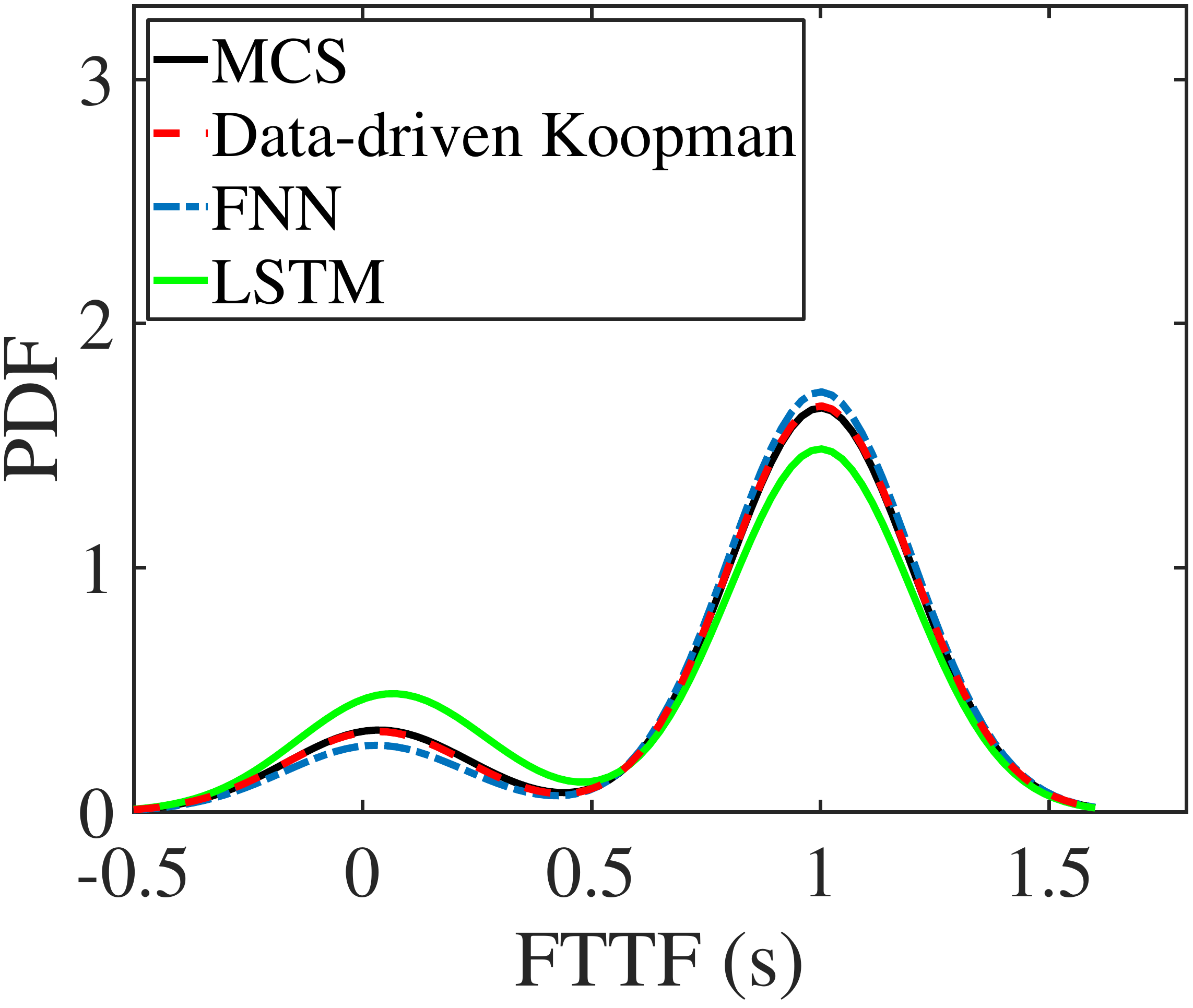}}
    \subfigure[]{
    \includegraphics[width=.4\textwidth]{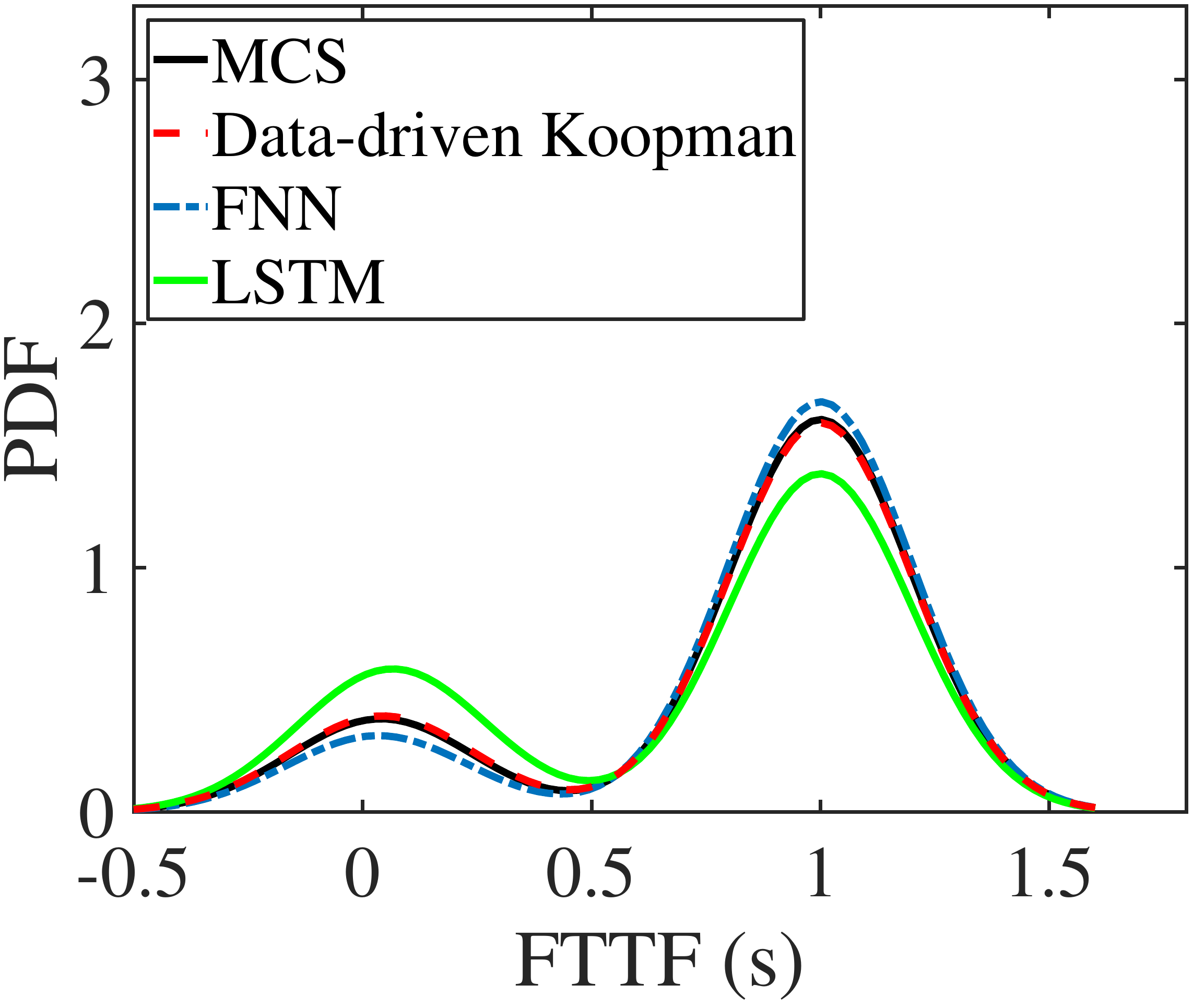}}
    \caption{PDF of failure time obtained by MCS, FNN, LSTM and proposed framework for the case of  uncertainties in the initial conditions, (a) with uniformly distributed samples of state variables generated in  $x\sim(0,20), y \sim(0,20)$ and $z \sim(-10,10)$ (b) with samples generated in $x\sim(-2,20), y \sim(-2,20)$ and $z \sim(-12,10))$}
    \label{fig:cased21}
\end{figure}

To exemplify the case of system with random initial conditions, probability density function of first passage failure time is computed for different sets of test samples.
For defining the first passage failure, the up-crossing thresholds are set as $e_h=[20,20,25]$. Note that unlike the case of duffing oscillator example, the threshold is defined for all the three state variables here.
\autoref{fig:cased21} depicts the  cases where PDF plots obtained for the test data generated in interval of $x\sim \mathcal{U}(0,20), y\sim \mathcal{U}(0,20)$ and $z\sim \mathcal{U}(-10,10)$. We observe that the proposed approach yields the best result followed by auto-regressive FNN.

\begin{figure}[!ht]
    \centering
    \subfigure[]{
    \includegraphics[width=.4\textwidth]{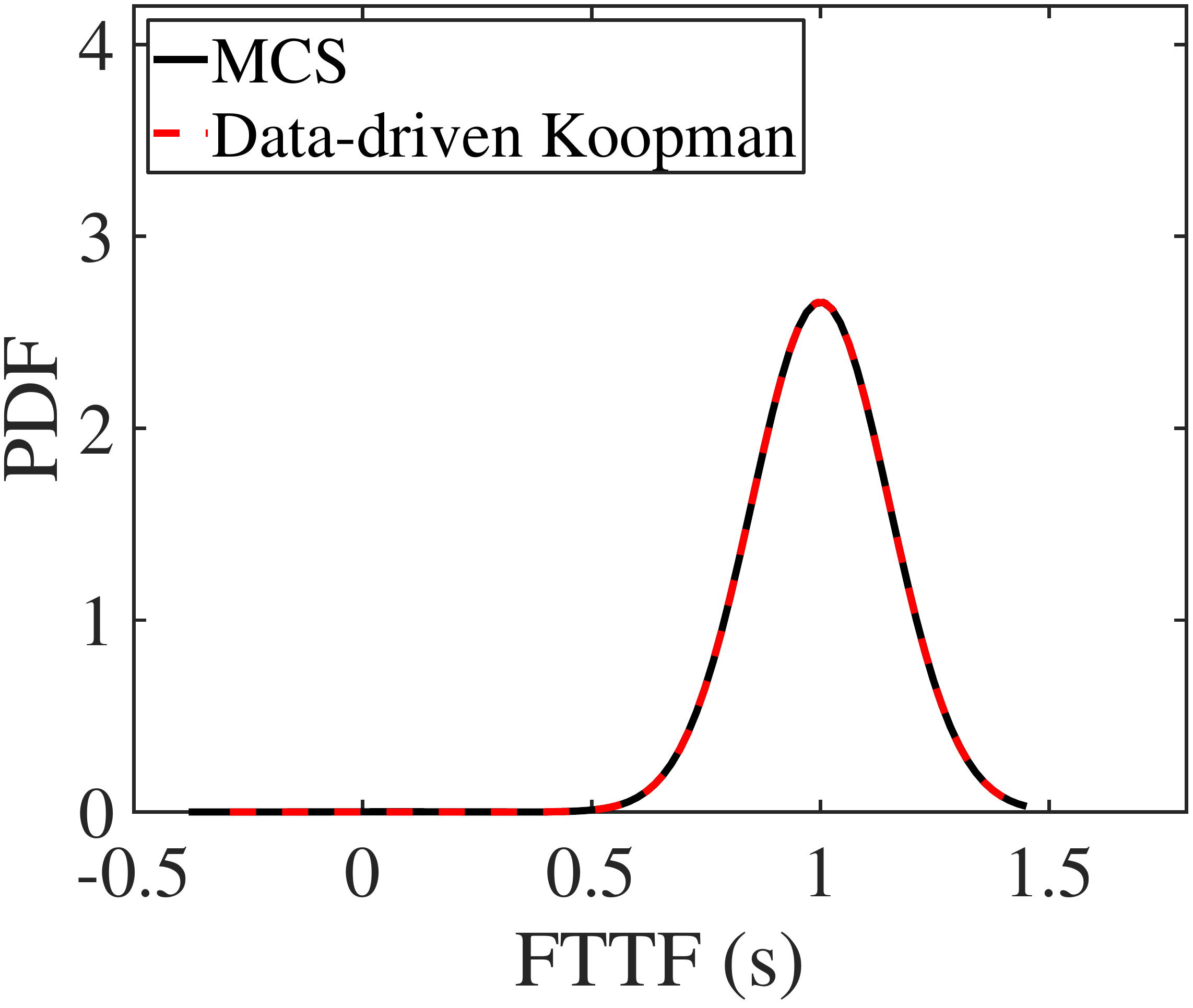}}
    \subfigure[]{
    \includegraphics[width=.4\textwidth]{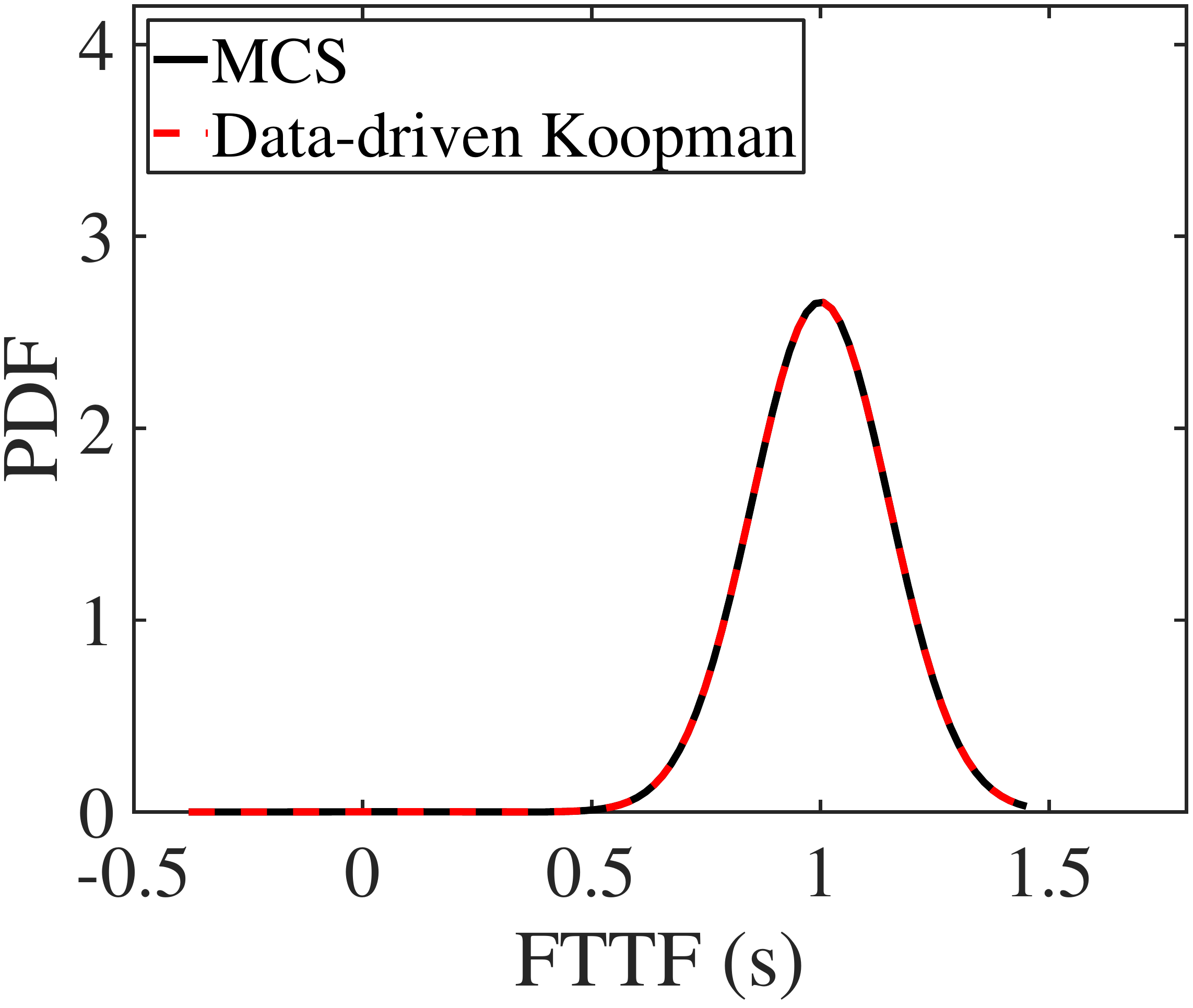}}
    \subfigure[]{
    \includegraphics[width=.4\textwidth]{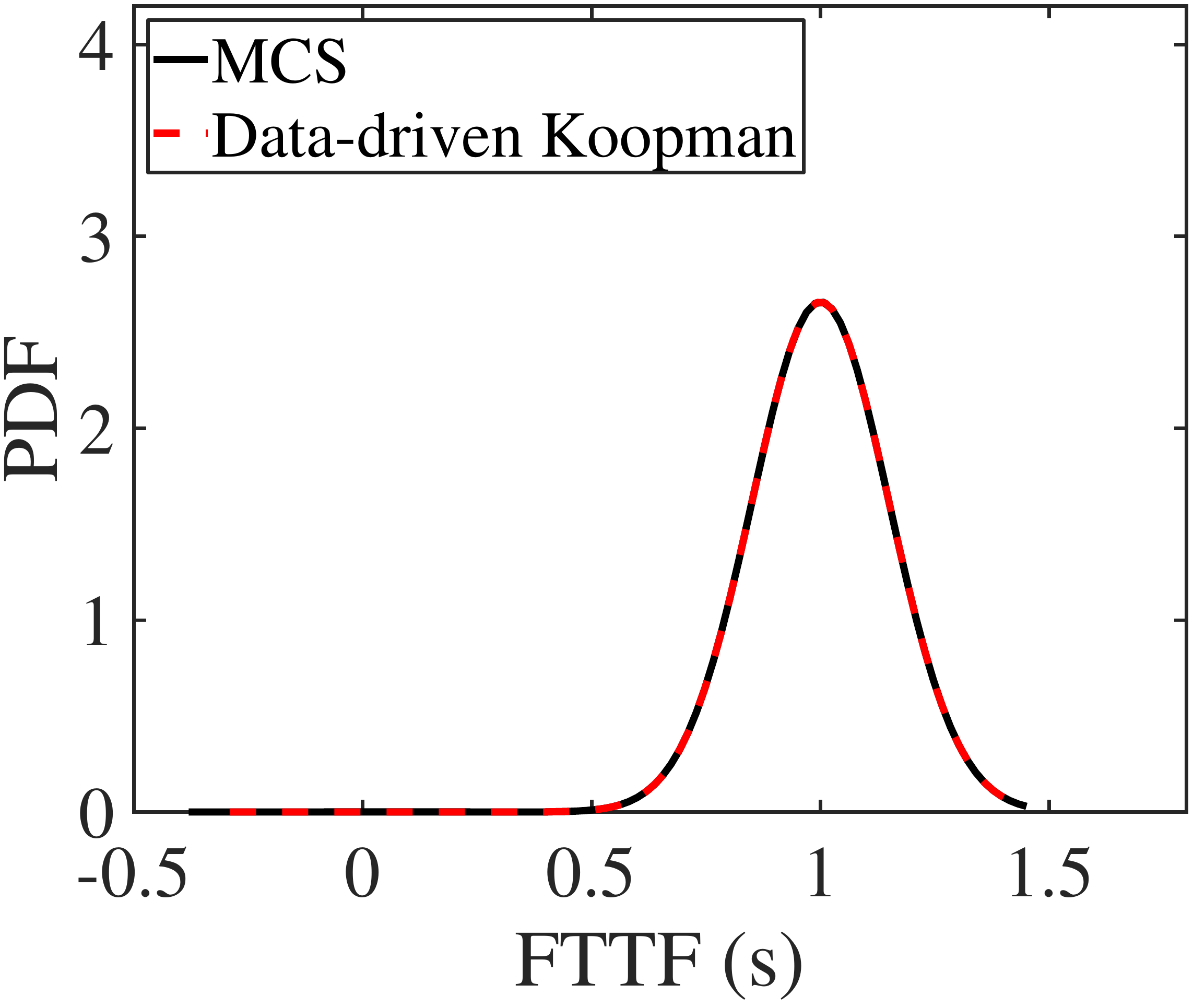}}
    \subfigure[]{
    \includegraphics[width=.4\textwidth]{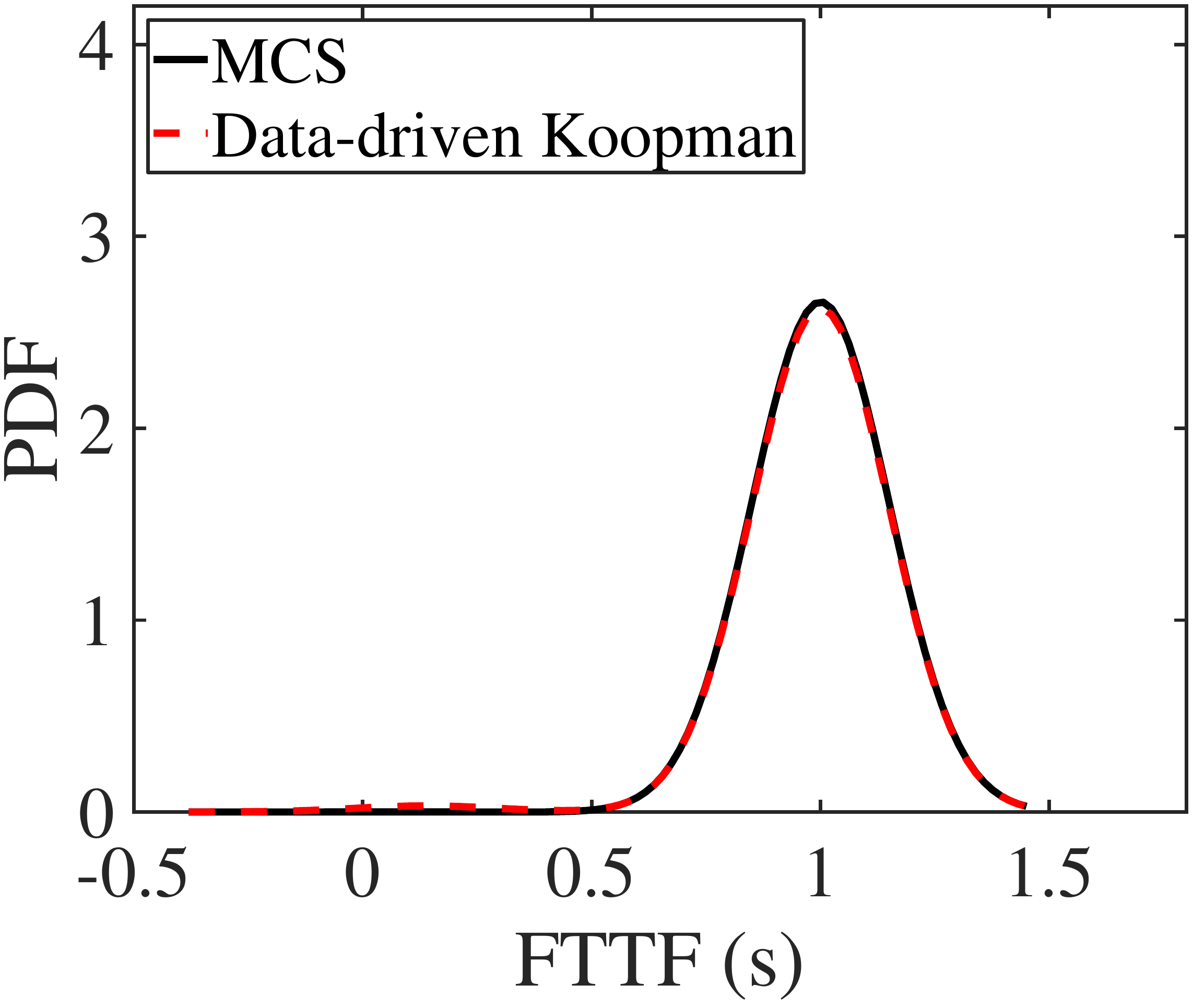}}
    \caption{PDF of failure time obtained by MCS and proposed framework for the case of  uncertainties in the system parameters, with samples of  parameters generated in $x\sim(0,20), y \sim(0,20)$ and $z \sim(-10,10)$ having distribution (a) Gaussian (b) lognormal, and with samples generated $x\sim(-2,20), y \sim(-2,20)$ and $z \sim(-12,10))$  having distribution (c) Gaussian (d) lognormal}
    \label{fig:cased21e}
\end{figure}

To demonstrate efficacy of proposed method for Out-of-distribution prediction, PDF plots of four different sets of test samples by varying range and distribution are shown in \autoref{fig:cased21e}.
The results obtained by the proposed framework achieves an excellent agreement with the MCS results in all the cases.  

Total failure probability and the reliability index are calculated by fixing the threshold, $e_h=[20,20,25]$. The results are showcased in \autoref{table3}. The proposed approach yields accurate results with a prediction error of $1.25\%$.
\begin{table}[ht!]
    \centering
    \caption{Results of first passage failure probability obtained using Proposed-framework for the case of Lorentz system with random initial conditions. $\beta_e$ indicates the reliability index obtained using MCS.}
    \label{table3}
\begin{tabular}{lcccc} 
\hline
\textbf{Method }& \textbf{Reliability Index} & \textbf{Failure Probability} & \textbf{$N_{s}$}&$\epsilon=\frac{|\beta_{e}-\beta|}{\beta_{e}}\times100$\\
\hline
MCS           &  0.966  &   0.170 &   $10^{4}$ &-\\ \hline
Koopman         &  0.954 &   0.167  &  1800 & 1.25\\ 
\hline 
\end{tabular}
\end{table}
\par
\begin{figure}[!ht]
    \centering
    \subfigure[]{
    \includegraphics[width=.4\textwidth]{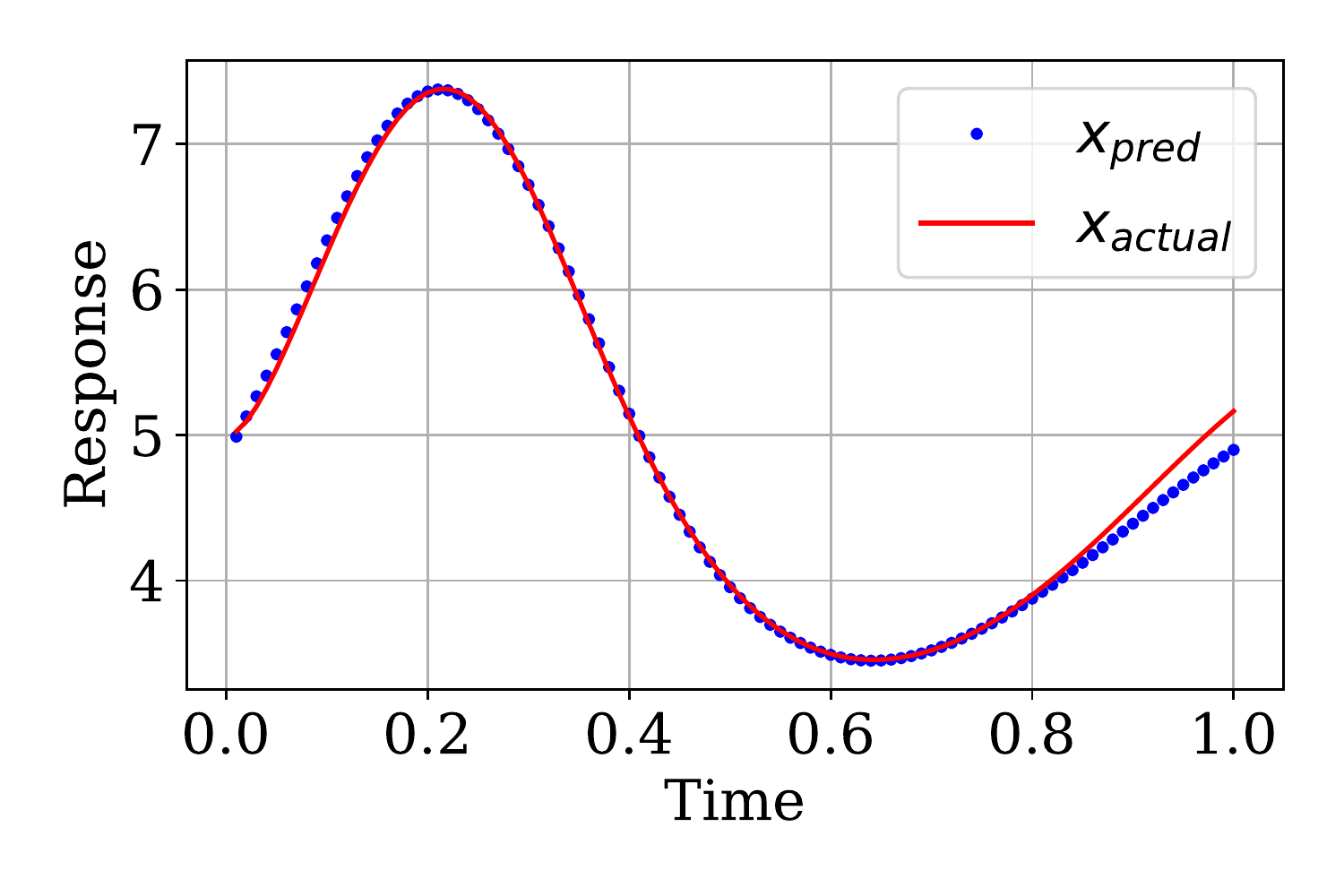}}
    \subfigure[]{
    \includegraphics[width=.4\textwidth]{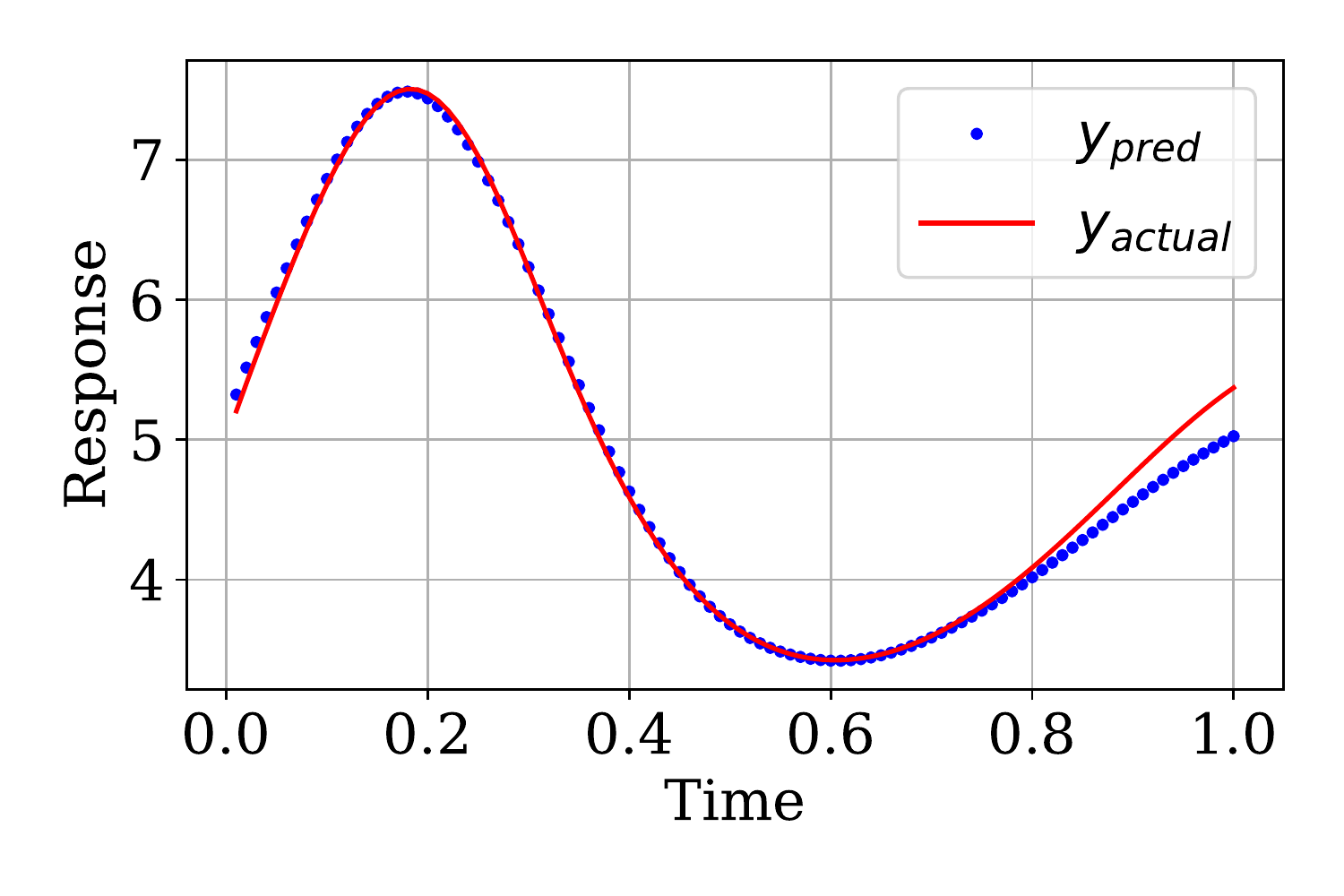}}
    \subfigure[]{
    \includegraphics[width=.4\textwidth]{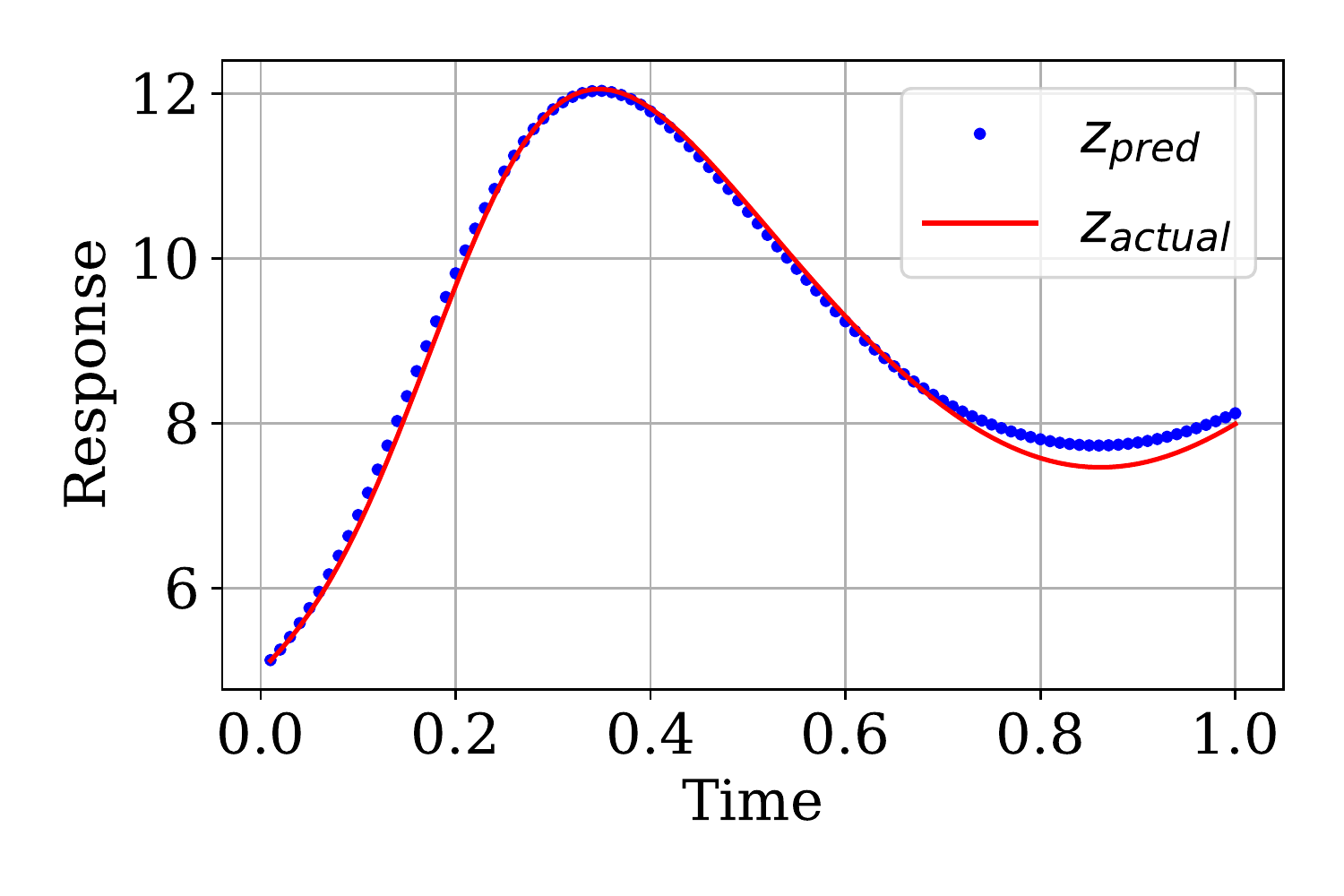}}
    \caption{State variables of Lorenz system predicted for random system parameters; (a) variation of $x$ with time, (b) variation of $y$ with time, (c) variation of $z$ with time}
    \label{fig:case22}
\end{figure}
\begin{figure}[!ht]
    \centering
    \subfigure[]{
    \includegraphics[width=.4\textwidth]{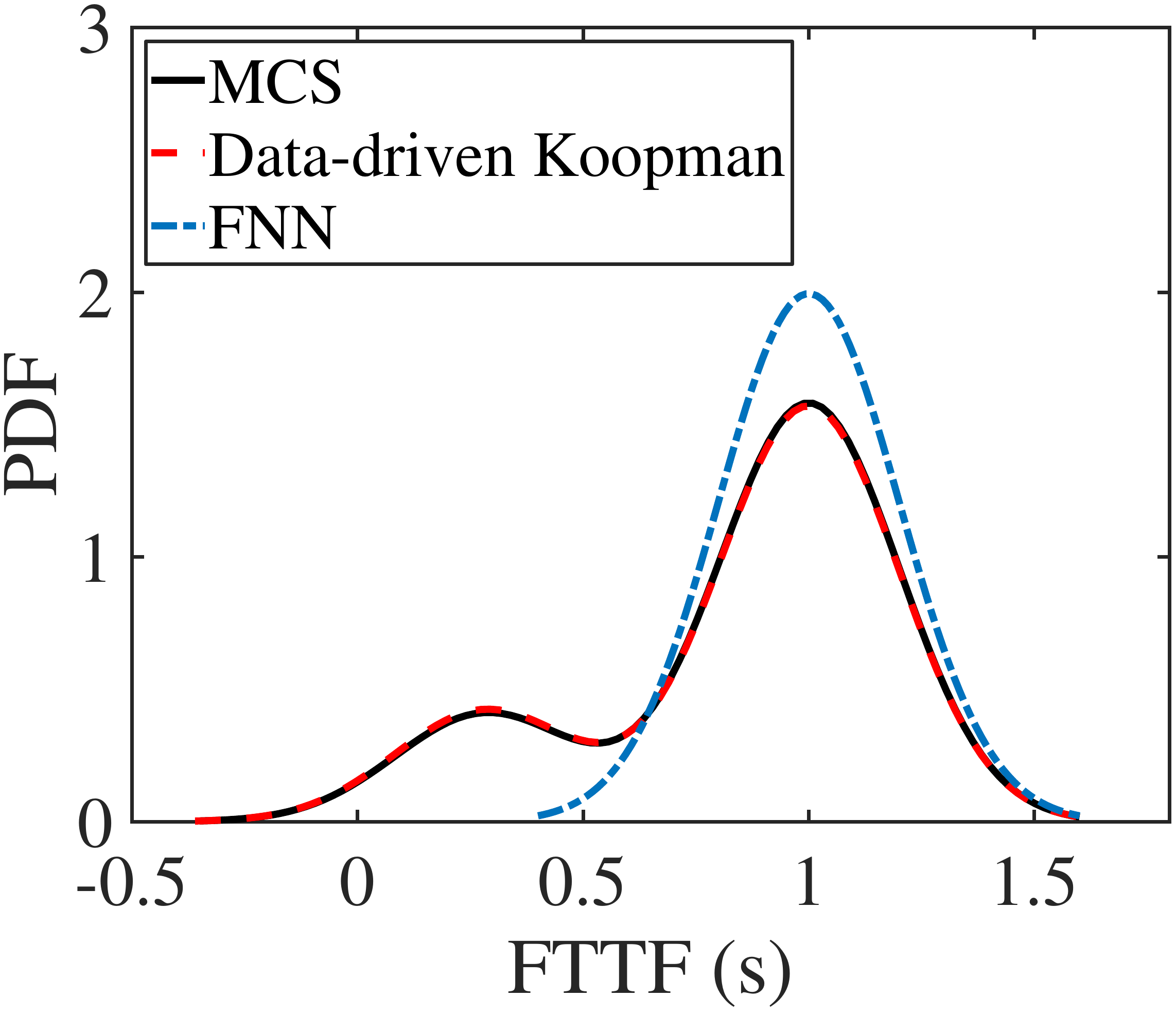}}
    \subfigure[]{
    \includegraphics[width=.4\textwidth]{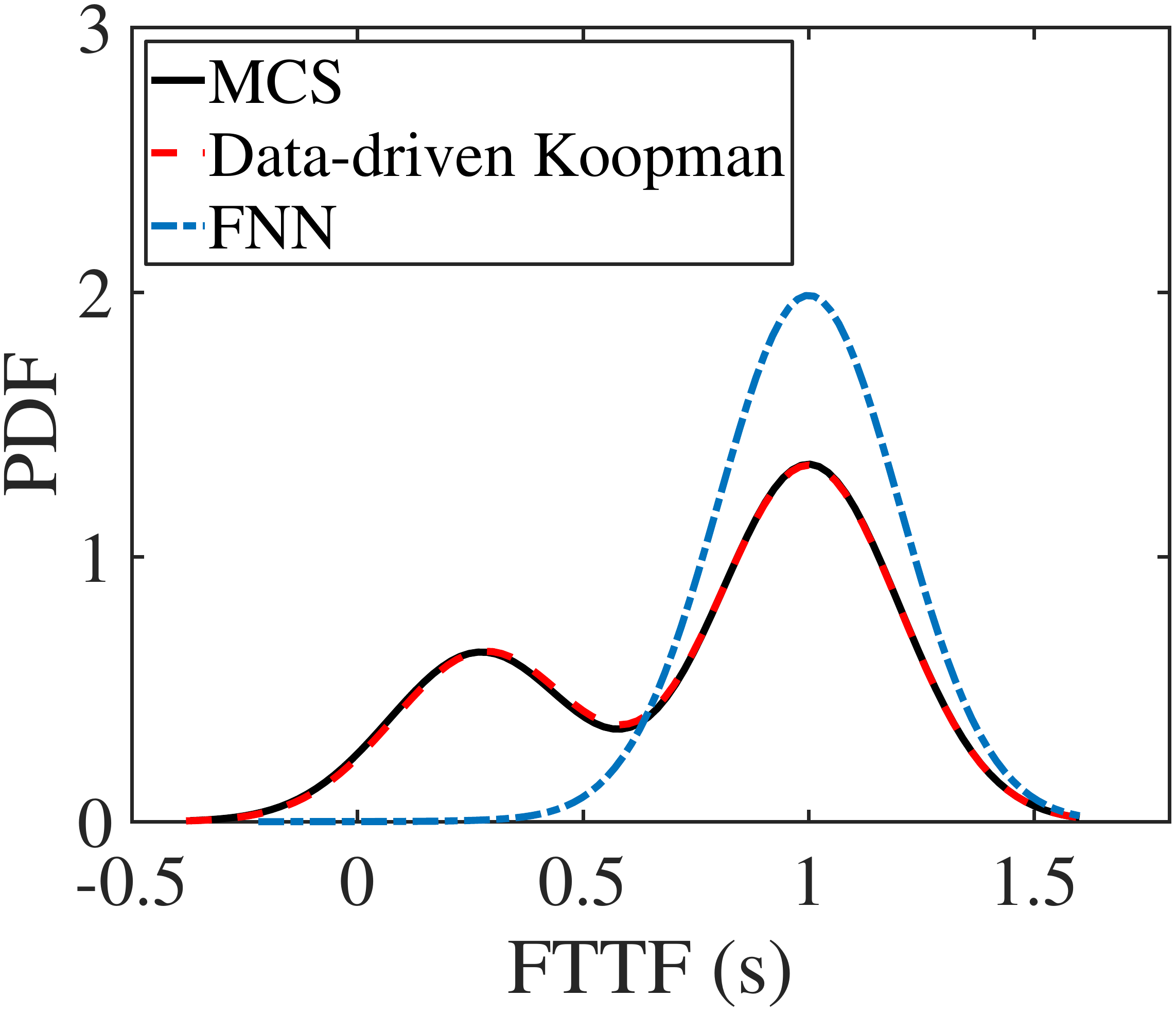}}
    \subfigure[]{
    \includegraphics[width=.4\textwidth]{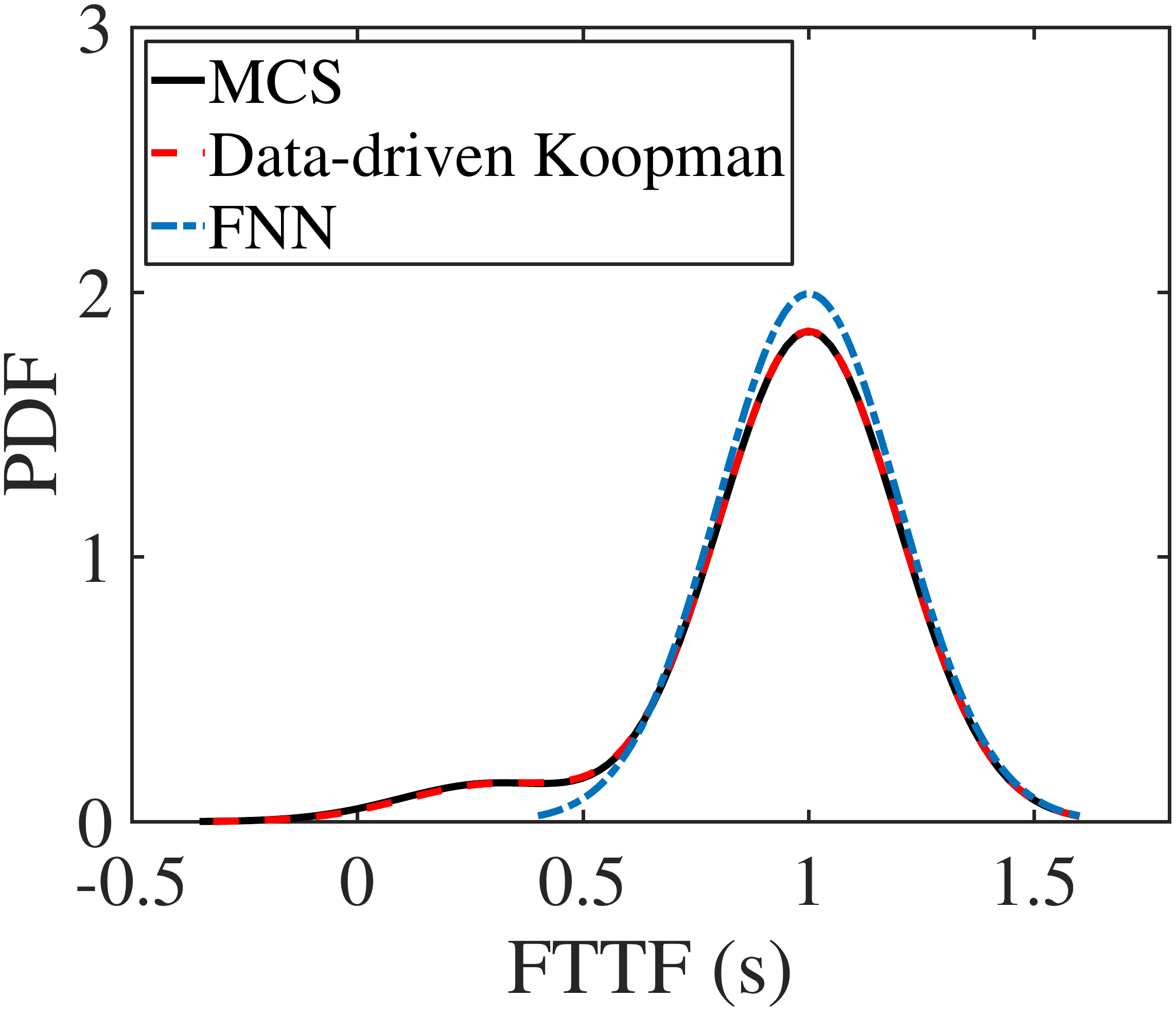}}
    \subfigure[]{
    \includegraphics[width=.4\textwidth]{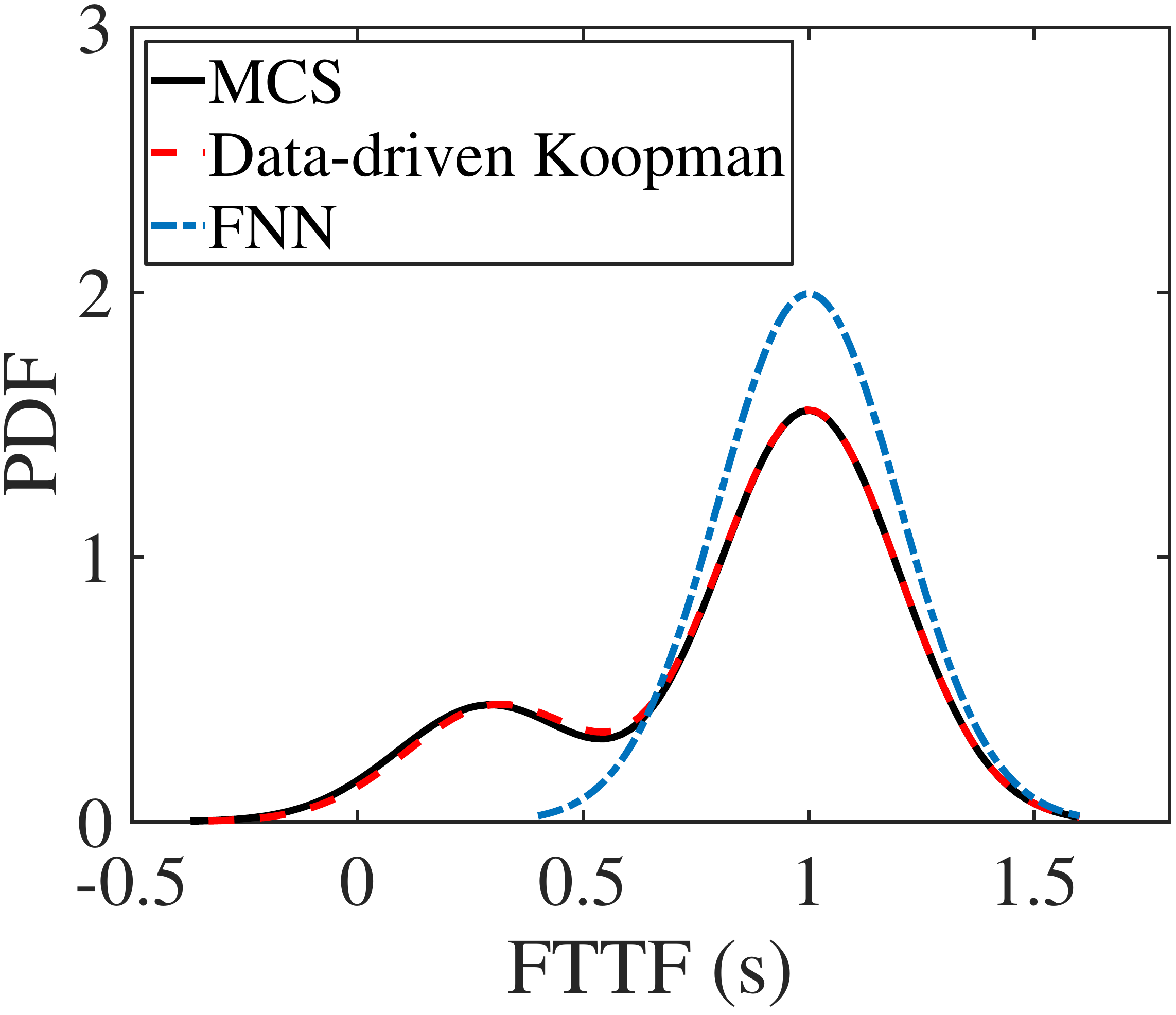}}
    \caption{PDF of failure time obtained by MCS and proposed framework for the case of  uncertainties in the system parameters, with samples of  parameters generated in $\sigma \sim(24,32), \rho \sim(8,12)$ and $\beta \sim(5/3,11/3)$ having distribution (a)uniform (b) Gaussian and with samples generated $\sigma \sim(23,33), \rho \sim(7.5,12.5)$ and $\beta \sim(5/3,14/3)$ having distribution (c) uniform (d) Gaussian}
    \label{fig:case22e}
\end{figure}

In regards to the variation of the system parameters, unlike the duffing oscillator, which has four system parameters, the Lorenz system has only three system parameters, i.e., $\sigma,\rho$ and $\beta$. For the case, uncertainties of the parameters are chosen as follows: $\sigma\sim \mathcal{U}(24,32), \rho \sim \mathcal{U}(8,12), \beta \sim \mathcal{U}(5/3,11/3)$. The architecture of the framework is similar to \autoref{fig:nnarch2} but modified to incorporate three Lorenz state variables and three system parameters. The model is trained with a data of $1800$ time series for 150 epochs and validated using data of $200$ time series. The response obtained for the validation data is shown in \autoref{fig:case22}.
The up-crossing limit are set to $e_h=[10,10,17]$, to define the first passage failure. By retaining the same threshold, first passage failure density plots obtained are shown in \autoref{fig:case22}. Here, the four sets of test samples of system parameters are generated from uniform and normal distributions such that  first two sets are in the support interval of $\sigma \sim (28,32), \rho \sim (8,12)$, and $\beta \sim (5/3,11/3)$  and other two sets are from the support interval of $\sigma \sim (27,33), \rho \sim (7.5,12.5)$, and $\beta \sim(5/3,14/3)$. The results are demonstrated in\autoref{fig:case22e}. As proposed approach outperforms the auto-regressive FNN and emulates the MCS results almost exactly. 
Fixing the limit state, $e_h=[10,10,17]$, we compute the total failure probability and reliability index in this case. The results presented in the \autoref{table4} shows the efficacy of the proposed method where the reliability index obtained with a prediction error of $3.7\%$
\begin{table}[ht!]
    \centering
    \caption{Results of first passage failure probability obtained using proposed-framework for the case of random system parameters . $\beta_e$ indicates the reliability index obtained using MCS.}
    \label{table4}
\begin{tabular}{lcccc} 
\hline
\textbf{Method }& \textbf{Reliability Index} & \textbf{Failure Probability} & \textbf{$N_{s}$}&$\epsilon=\frac{|\beta_{e}-\beta|}{\beta_{e}}\times100$\\
\hline
MCS           &  0.831  &   0.203 &   $10^{4}$ &-\\ \hline
Koopman         &  0.800 &   0.212  &  1800 & 3.7\\ 
\hline 
\end{tabular}
\end{table}
\subsection{Burger's Equation}
Burger's equation\cite{biazar2009exact,zhu2010numerical} is well studied partial differential equation, applied in various problems in fluid mechanics such as fluid turbulence, acoustic waves, and heat conduction. Here we consider Burger's equation for shock wave travelling in a viscous fluid medium as the third numerical example. The equation is given as:  
\begin{equation}\label{burger} 
    \begin{aligned}
    & \frac{\partial {u}}{\partial {t}}+ {u} \frac{\partial {u}}{\partial {x}}+ {v} \frac{\partial {u}}{\partial {y}} = \nu ({u} \frac{\partial^2 {u}}{\partial {x}^2}+{v}\frac{\partial^2 {u}}{\partial {y}^2})
    \\
    & \frac{\partial {v}}{\partial {t}}+ {u} \frac{\partial {v}}{\partial {x}}+ {v} \frac{\partial {v}}{\partial {y}}= \nu (\frac{\partial^2 {v}}{\partial {x}^2}+{v}\frac{\partial^2 {v}}{\partial {y}^2})  
    \end{aligned}
\end{equation}
where $u$ and $v$ are components of velocity in $x$ and $y$ directions respectively, $\nu$ represents the kinematic viscosity. For the given computational domain $D=\{(x,y):0 \leq x \leq 1, 0 \leq y \leq 1\}$, perturbation in the initial conditions \cite{zhu2010numerical} are given as:
\begin{equation}\label{eq3:initial}
    \begin{aligned}
    & u(x,y,0)=x+\alpha y\\
    & v(x,y,0)=x-\alpha y
    \end{aligned}
\end{equation}
Moreover, the kinematic viscosity is chosen to be constant for all the samples sets of simulation data, i.e. $\nu=0.01$.
\par
We have evaluated the efficacy of the framework to quantify the uncertainties, for the first two numerical examples, where the dynamical systems were described as ordinary differential equations (ODEs).  This example, on the other hand, is more challenging as the system here is described as Partial Differential Equations (PDEs). Unlike the previous examples, where the uncertainty in initial conditions and system parameters were used to evaluate the framework, here only the uncertainties in the initial conditions are considered. The initial conditions are obtained by varying  $\alpha$ in \autoref{eq3:initial} such that $\alpha \sim \mathcal{U}(0.6,1)$. Though there are different methods used to solve Burger's equation, to generate training, validation and testing data a finite-difference formulation is utilised here. The domain, $D$ is discretized into 64 grid points in the x and y dimensions for this purpose.
\par
\begin{figure}[!ht]
    \centering{
    \includegraphics[width=1\textwidth]{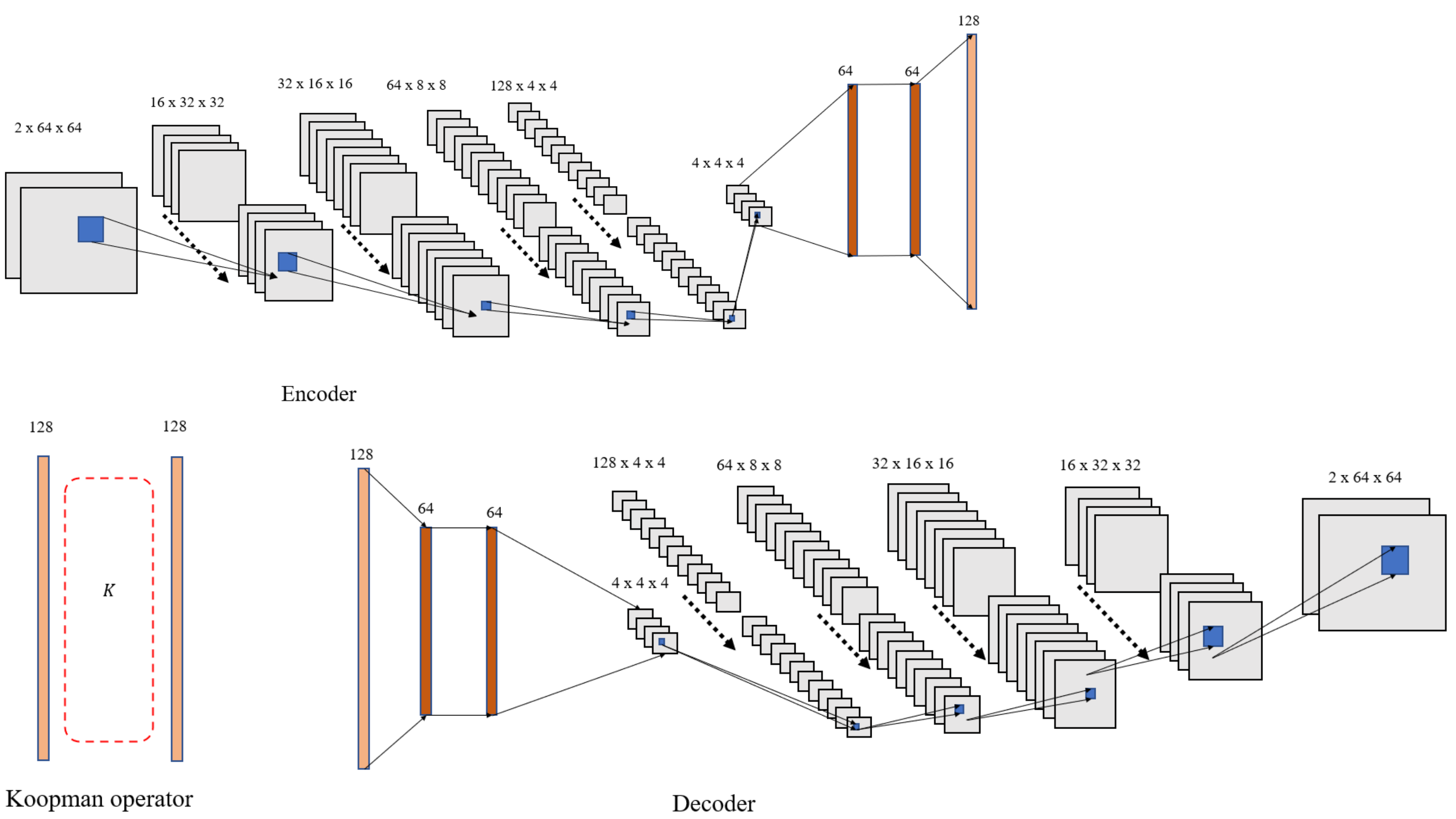}}
    \caption{Convolutional neural network based architecture of proposed Koopman framework for Burger's example with  uncertainty in the initial conditions}
    \label{fig:arch_conv4}
\end{figure}
\begin{figure}[!ht]
    \centering{
    \includegraphics[width=1\textwidth]{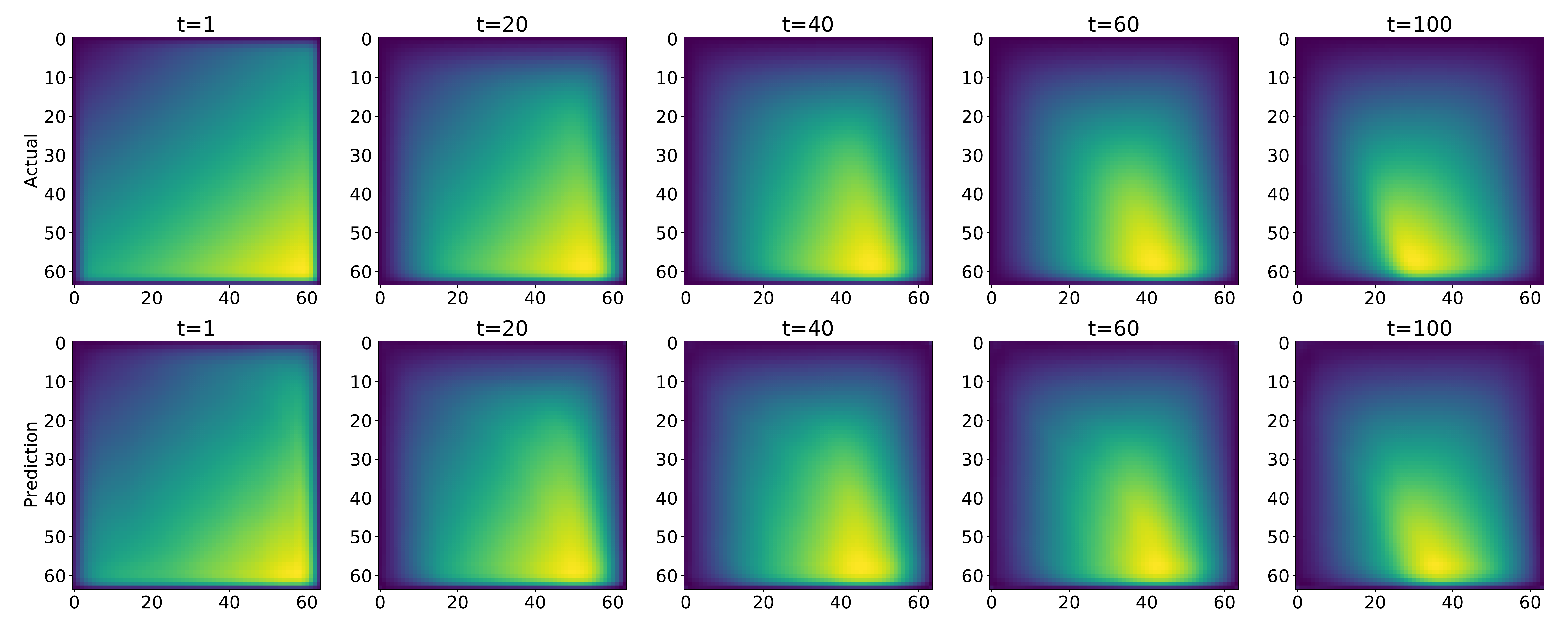}}
    \caption{Evolution of velocity field $u$ with time predicted by proposed framework}
    \label{fig:pred1}
\end{figure}
\begin{figure}[!ht]
    \centering{
    \includegraphics[width=1\textwidth]{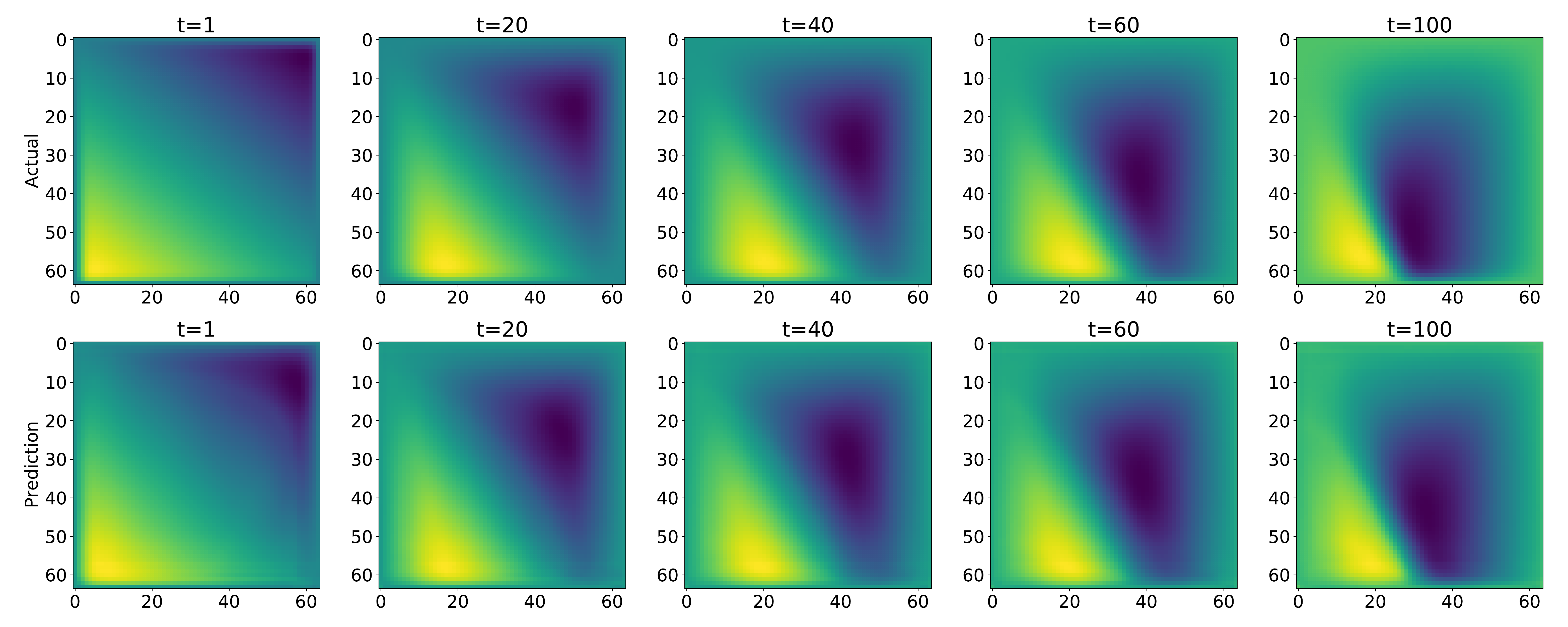}}
    \caption{Evolution of velocity field $v$ with time predicted by proposed framework}
    \label{fig:pred2}
\end{figure}
\begin{figure}[!ht]
    \centering
    \subfigure[]{
    \includegraphics[width=.4\textwidth]{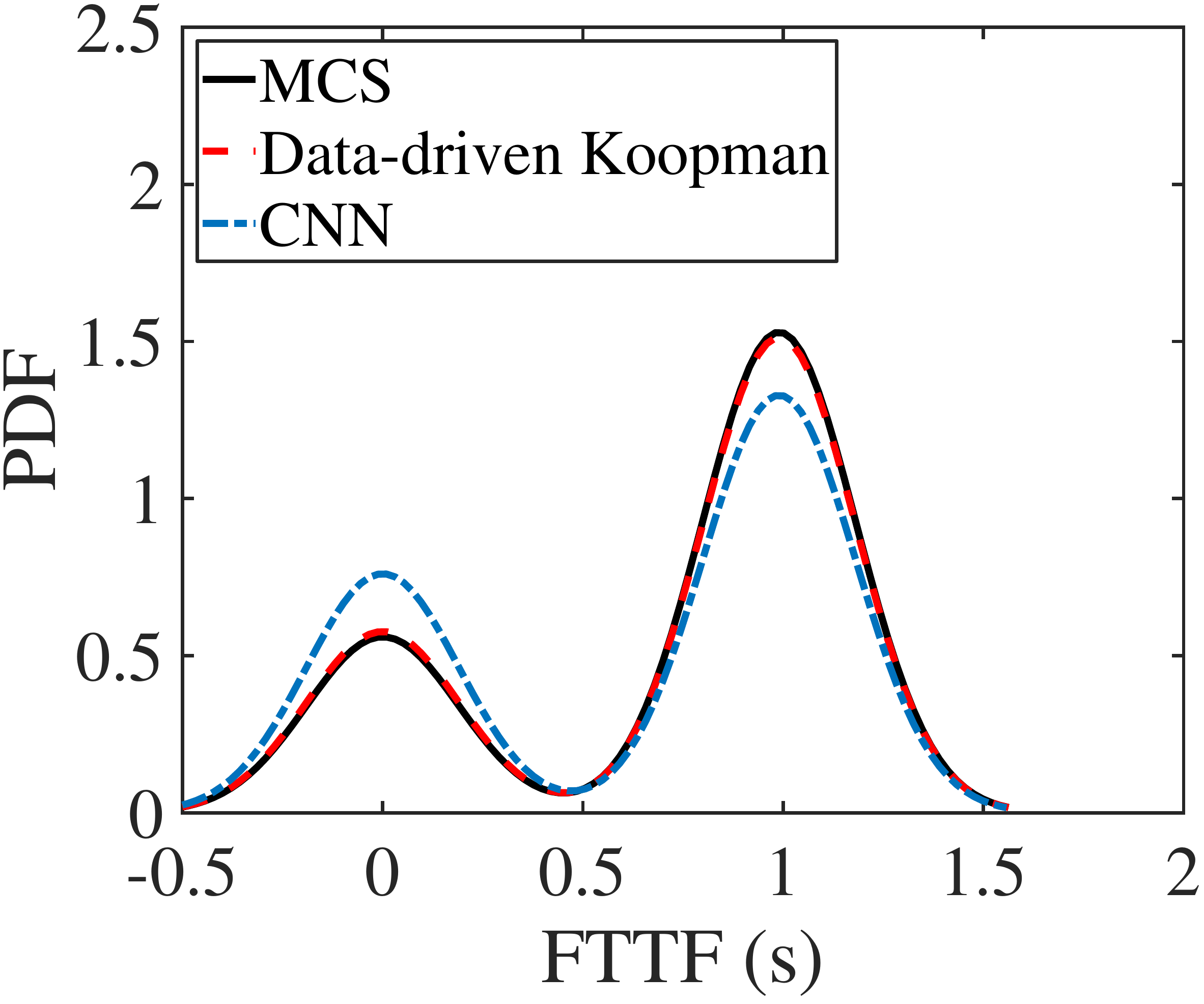}}
    \subfigure[]{
    \includegraphics[width=.4\textwidth]{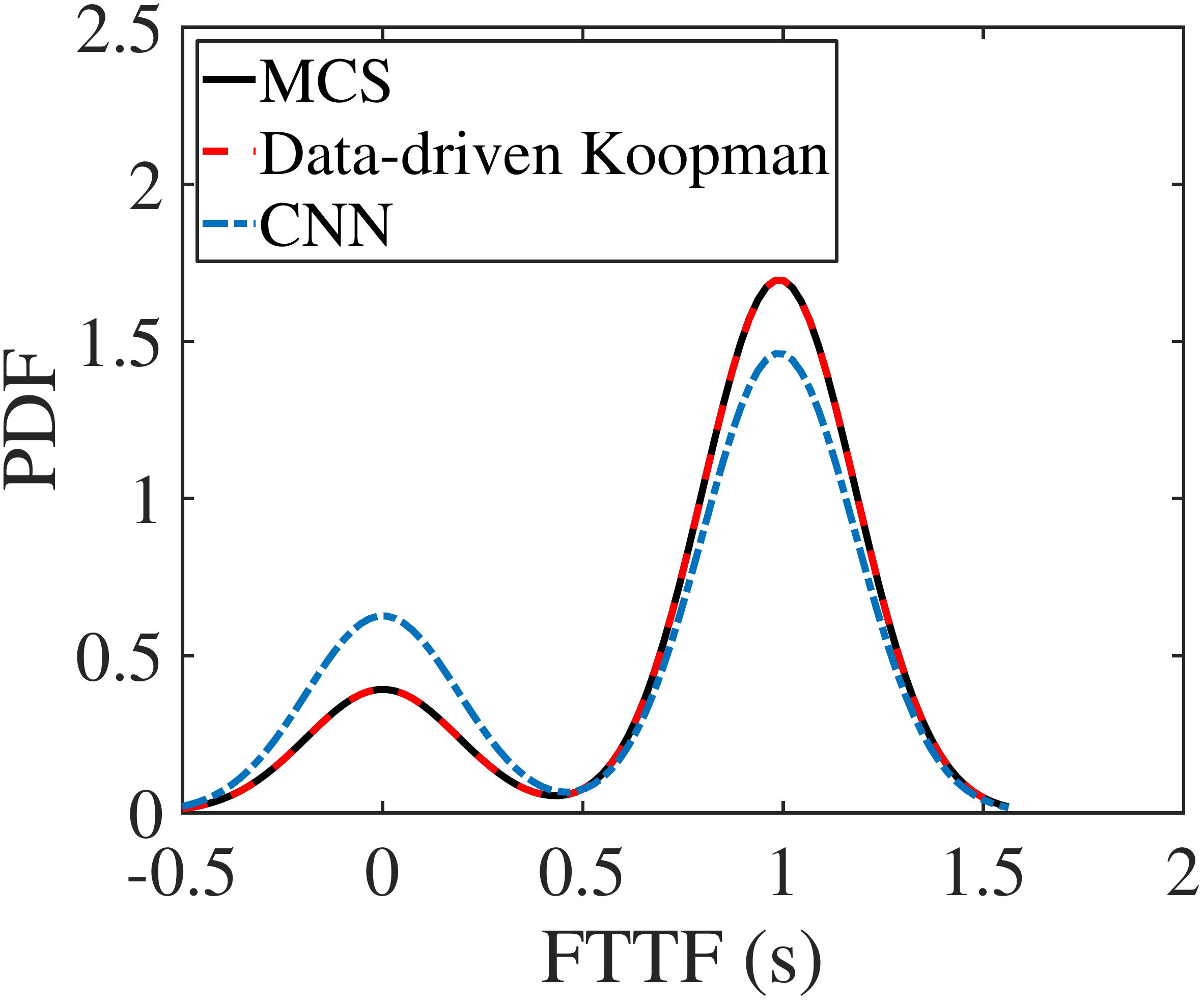}}
    \subfigure[]{
    \includegraphics[width=.4\textwidth]{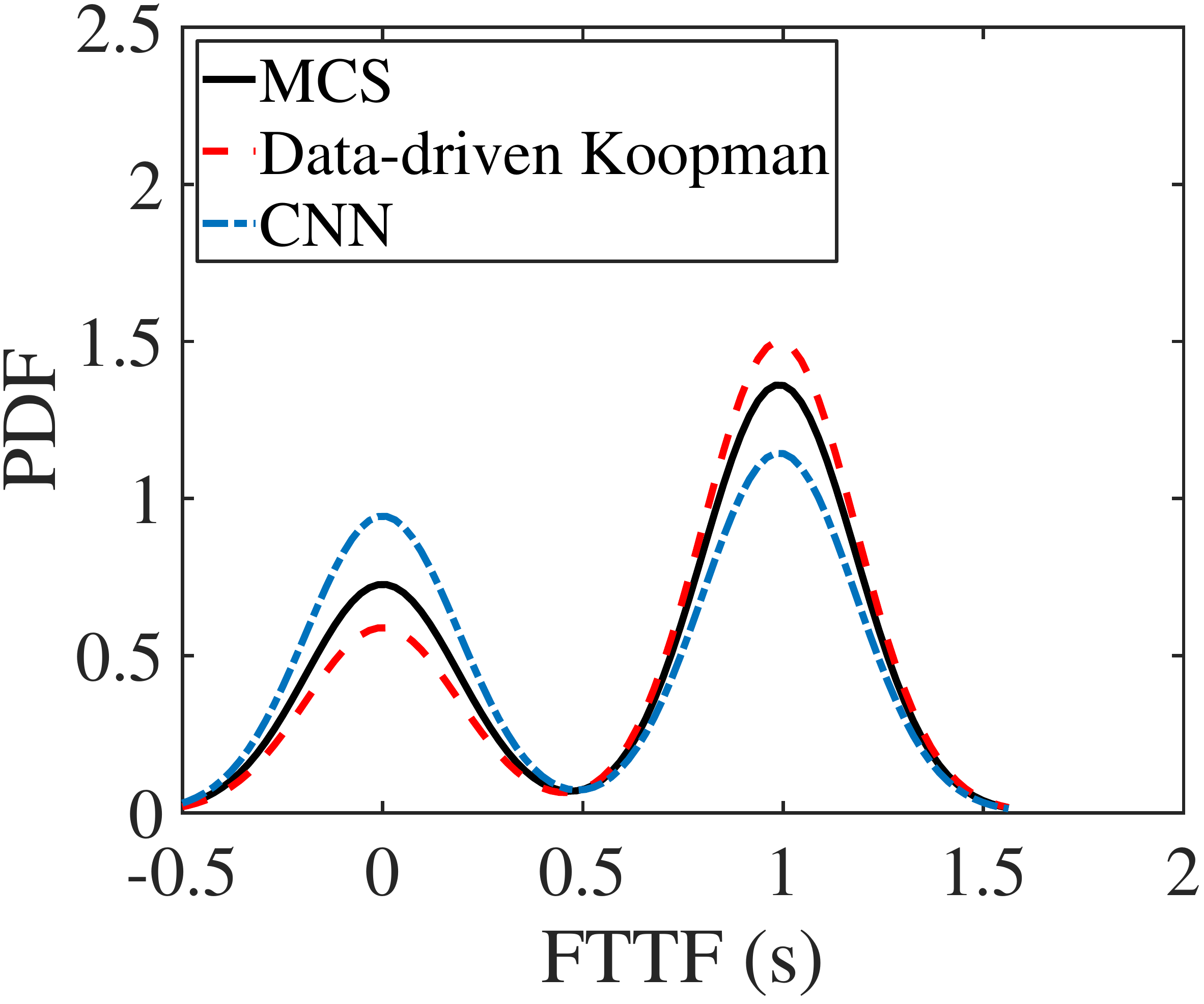}}
    \caption{PDF of failure time obtained by MCS and proposed framework, in Burger example having uncertainties in the initial conditions, for various limit states. The limit states of velocity fields $u$ and $v$ are (a)$[1.75,0.862]$ (b) $[1.78,0.862]$ (c) $[1.72,0.865]$}
    \label{fig:caselb21}
\end{figure}
\begin{figure}[!ht]
    \centering
    \subfigure[]{
    \includegraphics[width=.4\textwidth]{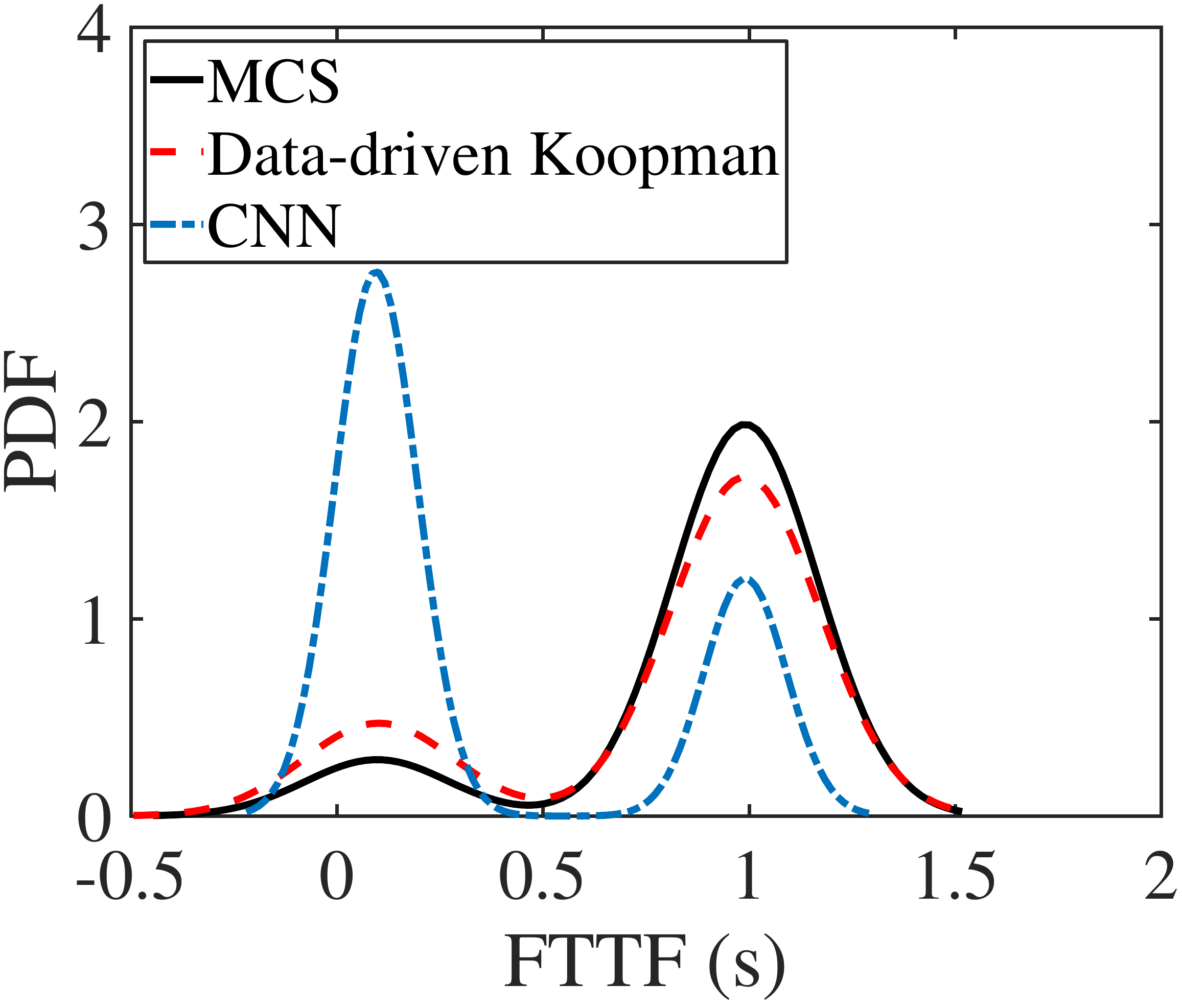}}
    \subfigure[]{
    \includegraphics[width=.4\textwidth]{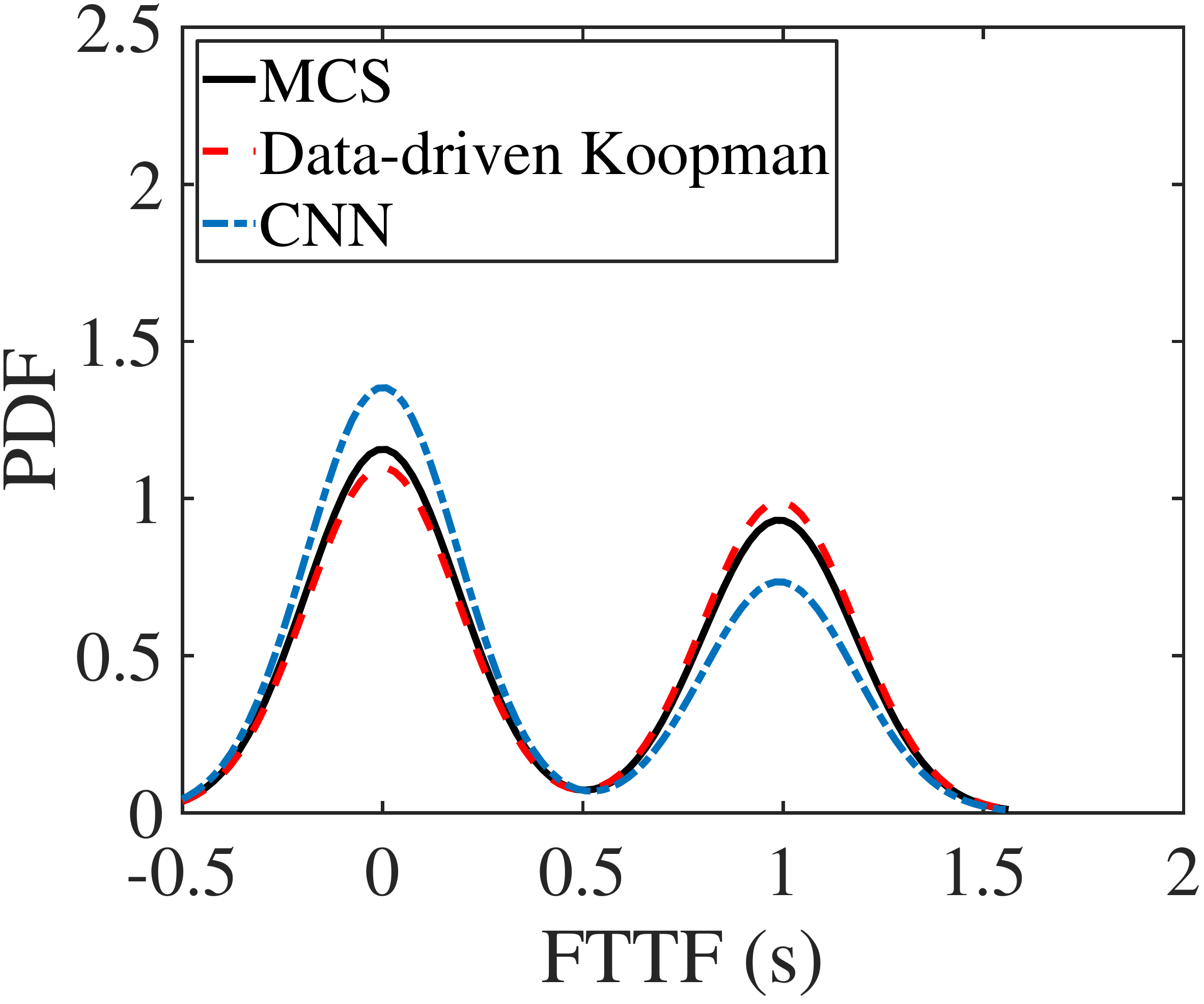}}
    \caption{PDF of failure time obtained by MCS and proposed framework, in Burger example having uncertainties in the initial conditions, for the limit states of  velocity fields $u$ and $v$ $[1.35,0.58]$  by fixing velocity field $u$ (a) at the grid point 20 (b) at grid point 40 along x axis}
    \label{fig:caselb22}
\end{figure}

The proposed framework architecture is depicted in \autoref{fig:arch_conv4}. Instead of a fully connected network, the framework uses the 2-D convolutional network (CNN). Encoder block contains 5 convolutional layers followed by three linear layers having ReLU activation function. The corresponding Koopman coordinates are supplied by the last linear layer of dimension 128. Finally, the decoder is designed complementary to the encoder block. The input of the of the encoder is comprised of two velocity components $\bm u$ and $\bm v$: $\{\bm u, \bm v\} \in \mathbb{R}^{2\times64 \times 64}$.
The training set and validation set consist of $80$ and $20$ time series data respectively, having 100-time steps with each time step of $10$millisecond. The trained model predicts system evolution based on initial conditions. The  model prediction of a random sample from the validation data after the  completing 100 epochs of training are shown in \autoref{fig:pred1} and \autoref{fig:pred2}. The model is able to accurately predict the time-marched states of the system.
The results of first passage failure time for various limit states are shown \autoref{fig:caselb21}. The PDF plots obtained from MCS, an auto-regressive CNN and proposed method are compared here. While the PDF of the proposed framework closely matches with the PDF obtained from MCS, PDF obtained from auto-regressive CNN clearly shows the contrast in comparison with the benchmark results.
Success of the proposed approach further demonstrated in \autoref{fig:caselb22}, where the PDF plots are obtained by fixing velocity field $u$ at grid point 20 and 40 along x axis. The PDF plots show the close agreement of results of the proposed Koopman framework with the MCS. On the other hand, the results obtained by CNN is inconsistent in comparison with the benchmark results. 
\section{Conclusions}\label{sec:Conclusions}
In this work, we have proposed a novel approach that exploits Koopman operator to evaluate time-dependent reliability analysis of dynamical systems. However, Koopman operators are not known a-priori. To address this issue, we propose a deep learning architecture that first learn the Koopman operator and then uses the same for predicting the dynamical. Unlike the popular data driven auto-regressive and forecasting models, the proposed approach yields accurate results even in the presence of uncertainty and generalizes to other distributions; this renders the proposed framework suitable for solving time-dependent reliability analysis problems.
\par
The proposed framework is designed to evaluate the first passage failure probability of the system in the presence of uncertainties in the initial conditions as well as the system parameters. In the present work, we have considered the two cases separately, though this can be also extended to the case where uncertainties in both initial conditions and system parameters exist concomitantly. Three numerical examples of dynamical systems described by ODES and PDE have been solved using the proposed method. The results are compared with the state-of-the-art surrogate methods and the benchmark MCS results. It is observed that the proposed approach outperforms the other methods, yielding accurate results for first passage failure time. The superiority of the proposed approach is particularly visible in case of uncertain parameters. Moreover the fact that the proposed approach yield accurate estimate of first passage failure time for chaotic Lorenz system is especially noteworthy. On the other hand, the scalability of the proposed approach to systems governed by PDE is evident from the two dimensional Burger's equation illustrated in this paper.  
\section*{Acknowledgements}
NN acknowledges the support received from Ministry of Education in the form of Prime Ministers Research Fellowship grand. SC acknowledges the financial support of Science and Engineering Research Board (SERB) via grant no. SRG/2021/000467.

% \bibliographystyle{ieeetr}
% \bibliography{references}  %%% Uncomment this line and comment out the ``thebibliography'' section below to use the external .bib file (using bibtex).

\end{document}